\RequirePackage[hyphens]{url}
\documentclass[10pt,journal,compsoc]{IEEEtran}
\pdfoutput=1

\ifCLASSOPTIONcompsoc
  \usepackage[nocompress]{cite}
\else
  \usepackage{cite}
\fi
\PassOptionsToPackage{svgnames}{xcolor}
\usepackage{svg}
\usepackage{ragged2e}
\usepackage{xspace} 
\usepackage{amsmath}
\usepackage{enumitem}
\usepackage{amsfonts}
\usepackage{hyperref}
\usepackage[nopar]{lipsum}
\usepackage{soul}
\usepackage{makecell}
\usepackage{svg}
\usepackage{wrapfig}
\usepackage{scalerel}
\usepackage{amsmath, nccmath}
\usepackage{floatrow}
\usepackage{dblfloatfix}
\AtBeginEnvironment{align*}{\useshortskip}
\definecolor{highlight}{cmyk}{0, 0.7808, 0.4429, 0.1412}

\usepackage{array}
\newcommand{\highlight}[1]{{{#1}}}

\newcommand{\name}{Deep~Umbra\xspace}
\newcommand{\dataset}{Global Shadow Dataset\xspace}
\renewcommand{\paragraph}[1]{\noindent\textbf{{#1}}.}

\begin{document}
%
\title{Deep Umbra: A Generative Approach for Sunlight Access Computation in Urban Spaces}
%
%
%
%

\author{Kazi~Shahrukh~Omar,
        Gustavo~Moreira,
        Daniel~Hodczak,\\
        Maryam~Hosseini,
        Nicola Colaninno,
        Marcos~Lage,
        and~Fabio~Miranda
\IEEEcompsocitemizethanks{\IEEEcompsocthanksitem K. Omar, G. Moreira, D. Hodczak, and F. Miranda are with the University of Illinois Chicago.\protect\\
E-mail: \{komar3, gmorei3, dhodcz2, fabiom\}@uic.edu
\IEEEcompsocthanksitem M. Hosseini is with the Massachusetts Institute of Technology.\protect\\
E-mail: maryamh@mit.edu
\IEEEcompsocthanksitem N. Colaninno is with the Polytechnic University of Milan.\protect\\
E-mail: nicola.colaninno@polimi.it
\IEEEcompsocthanksitem M. Lage is with the Universidade Federal Fluminense.\protect\\
E-mail: mlage@ic.uff.br
}

\thanks{Manuscript accepted February, 2024.}}

%
%

\markboth{IEEE Transactions on Big Data,~Vol.~X, No.~X, February~2024}%
{Shell \MakeLowercase{\textit{et al.}}: Bare Demo of IEEEtran.cls for Computer Society Journals}
%



\IEEEtitleabstractindextext{%
\begin{abstract}
\justifying
Sunlight and shadow play critical roles in how urban spaces are utilized, thrive, and grow. While access to sunlight is essential to the success of urban environments, shadows can provide shaded places to stay during the hot seasons, mitigate heat island effect, and increase pedestrian comfort levels. Properly quantifying sunlight access and shadows in large urban environments is key in tackling some of the important challenges facing cities today. In this paper, we propose \name, a novel computational framework that enables the quantification of sunlight access and shadows at a global scale. Our framework is based on a conditional generative adversarial network that considers the physical form of cities to compute high-resolution spatial information of accumulated sunlight access for the different seasons of the year. We use data from seven different cities to train our model, and show, through an extensive set of experiments, its low overall RMSE (below 0.1) as well as its extensibility to cities that were not part of the training set. Additionally, we contribute a set of case studies and a comprehensive dataset with sunlight access information for more than 100 cities across six continents of the world. \name is available at \href{http://urbantk.org/shadows}{urbantk.org/shadows}.
\end{abstract}

\begin{IEEEkeywords}
Urban computing, Urban analytics, Sunlight access, Shadow, Generative adversarial networks.
\end{IEEEkeywords}}

\maketitle

\IEEEdisplaynontitleabstractindextext

%
\IEEEpeerreviewmaketitle

\section{Introduction}
\label{sec:introduction}

The urban population surpassed the rural one in 2007, and by 2030 it is expected to represent 60\% of the global population and 68\% by 2050~\cite{UN2019}.
This leads to rapid urban development around the world, requiring sustainable solutions, social equity, and urban regeneration~\cite{ding2015inclusive}.
Urban growth implies two types of development: horizontal (through the use of land for new infrastructure, e.g., buildings, parks) and vertical (through densification and construction of taller buildings).
Denser cities, however, can impact urban environmental conditions and quality of life, influencing solar access and shadows in outdoor spaces and recreational areas~\cite{jasonMBarr19}.
The appropriate management of shadows then plays a vital role in maintaining the citizens' quality of life.
%
%
As previously highlighted~\cite{miranda2018shadow}, it is crucial to have efficient methods that allow stakeholders to (1) comprehensively analyze the shadow and sunlight access in a region, and (2) democratize the planning process, allowing communities and the general audience to be aware of the potential disruptions of their \emph{right to light}.

In previous years, this problem has been tackled from different angles and by different domains~\cite{compagnon2004solar, kruger2011impact, jiang2017sunchase, vulkan2018modeling}. However, such studies were oftentimes not scalable beyond a handful of cities (or even just neighborhoods), only quantifying shadows over a limited set of timeframes. 
At the core of our research is the understanding that increasing awareness about this pressing issue and supporting sustainable urban development efforts require new methods and datasets that enable the \emph{comprehensive} and \emph{fast} analysis of shadows in cities around the world over long periods of time. 
By comprehensive, we argue that it is not enough for analyses to be constrained to a handful of timestamps, but there is a need to \emph{accumulate} shadows and generate data inventories with the percentage of sunlight access at a given point over a long period of time -- e.g., a given point is under shadow for 70\% of time from sunrise to sunset.
By fast, we argue that the aforementioned accumulation computations should be performed as fast as possible, as to allow for large-scale interactive and exploratory analyses.
Previous approaches tackled these challenges by ray tracing to accumulate shadows over time~\cite{miranda2018shadow}, but still with performance results that limit the scale and scope of analyses. Popular GIS tools, for example, take hours to compute the shadow / sunlight access data for a single large city, such as New~York~City~(NYC). Such performance limits the scale and scope of analyses by urban experts.


This research is supported by the growing availability of urban data describing the built environment (in our case, building heights) as well as recent developments in machine and deep learning that enable image-to-image translation tasks. 
With sufficient training data, we frame our problem around tile images with height information and propose a \emph{height tile}$\rightarrow$\emph{accumulated shadow tile} translation task. We also take into account a city's latitude and time of the year, as these two factors greatly impact the shape of shadows.
%
With this in mind, we propose \name, a novel computational framework that uses a conditional generative adversarial network (GAN) for the quantification of sunlight access and shadows over long periods of time and at a global scale. Our framework considers the physical form of cities to compute high-resolution spatial information of \emph{accumulated} sunlight access for the different seasons of the year.
As shadows also depend on the latitude of a city, a key aspect of our research is to ensure transferability to varying locations around the world. We enforce this by training our model in urban areas from different continents and with disparate physical forms. 
Our experiments show how this approach leads to good results in cities outside the training set.
Moreover, in order to enable research in domains interested in the impact of shadows and sunlight access in public spaces, we make our outcomes publicly available at \href{http://urbantk.org/shadows}{urbantk.org/shadows}.

\paragraph{Contributions} In this paper, we take a step towards enabling the fast accumulation of shadows over time using a conditional generative adversarial network.
We first describe how shadow accumulation can be quantified over geographical regions. We then introduce a conditional GAN framework to efficiently accumulate shadows over time. This approach is then used to compute accumulated shadow information for over 100 cities in the world.
We also report two case studies performed by urban experts (co-authors of the paper) using the data.
These studies highlight the opportunities created by such data, not only for fine-grained analyses across neighborhoods of a city, but also large-scale exploratory analyses across different cities.
%
%
Our contributions can then be summarized as follows:
\begin{itemize}
    \item We propose \name, a computational framework for the quantification of sunlight access and shadows over long periods of time with a speed up of 6$\times$ against the state of the art~\cite{miranda2018shadow}.
    \item We introduce the \dataset, a comprehensive dataset with the accumulated shadow information for over 100 cities in 6 continents.
    \item We show experimental evaluation demonstrating the accuracy of our framework, including its transferability to cities outside of the training set.
    \item In collaboration with urban experts, we demonstrate the usefulness of \name and the \dataset through a set of case studies evaluating sunlight access over street networks and parks.
\end{itemize}

This paper is organized as follows: In Section~\ref{sec:related} we review important related work; Section~\ref{sec:background} briefly reviews information regarding the accumulation of shadows; Section~\ref{sec:methodology} and Section~\ref{sec:data} present \name and the accumulation shadow data, respectively; In Section~\ref{sec:experiments} we present a detailed evaluation of \name; Section~\ref{sec:dataset} presents the \dataset; Section~\ref{sec:cases} highlights two case studies. Finally, Section~\ref{sec:conclusion} concludes our paper and presents future research directions.
\begin{figure*}[t]
  \centering
  \includegraphics[width=1\linewidth]{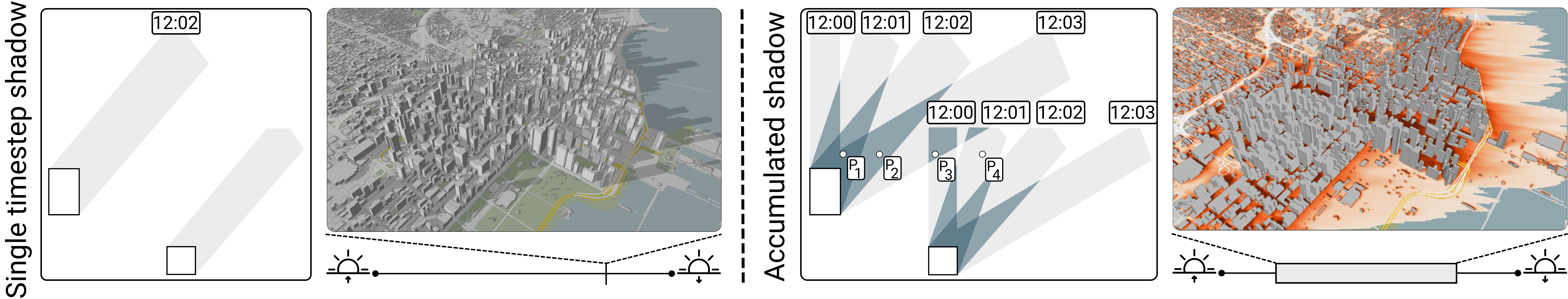}
  \caption{Left: Example of a single timestep shadow. Right: accumulated shadows. By accumulating shadows between a time range, we can comprehensively analyze the impact of buildings on the sunlight access of public spaces. In the illustration, $P_1$ is 100\% of the time under shadow, $P_2$ and $P_3$ 50\%, and $P_4$ 25\%. The rightmost image shows the resulting accumulation when considering a time range between 10~AM and 4~PM.}
  \label{fig:example}
\end{figure*}

\section{Related Work}
\label{sec:related}

In this section, we review previous works highlighting the significance of shadows to different domains, shadow computation techniques and GAN-assisted urban analytics.

\paragraph{Study of sunlight access and shadows}  Due to population growth, urban constructions have increasingly started to propagate vertically, creating the problem of uneven solar access in cities. The concept named ``Solar Envelope'' was first introduced by Knowles~\cite{knowles1980solar}. His idea was to combine urban building design with the impact of shadow and solar accessibility. Studies have extended upon this idea of solar envelope by proposing vertical limits and distance regulations for sites under construction to ensure sunlight access for dwellings in cities~\cite{de2019novel, alzoubi2010low, pereira2001methodology}. In addition, design layouts for buildings have been proposed in several works aiming at utilizing solar potential~\cite{hachem2013solar, vartholomaios2015residential, kanters2016planning}.

Sunlight access and impact of shadow are key factors for a comfortable microclimate at street level for pedestrians. Studies have extensively discussed the contribution of the street network design and solar orientation to outdoor shadow impact and thermal comfort levels~\cite{andreou2014effect, deng2020impact}. In major cities, the surge of high-rise buildings is also sinking the public spaces into deepening shadow, calling into question the protection of sunlight in open spaces. Different studies have been conducted to properly measure sunlight access in parks and other public recreational areas \cite{seelye2017building, costamagna2019livability, zhu2019solar}.

By making available both our model as well as a comprehensive global dataset with shadow information, our contributions can greatly increase the spatial coverage of previous works and offer the first steps towards a more comprehensive understanding of this pressing problem.

\paragraph{Shadow computation techniques} 
At the core of our paper is determining whether a point is under shadow or not for one or more timestamps. This problem has been comprehensively explored in computer graphics~\cite{eisemann2011real}.
In general, real-time shadow computation can be performed through \emph{shadow mapping}, an image-based approach in which the 3D scene is projected onto a 2D plane situated between the scene and the light source.
%
%
%
In contrast, ray tracing is a robust and popular approach in which computation is carried out by tracing rays from the light source to the objects~\cite{nah2014sato}.
%
%
More recently, deep learning approaches have been proposed for the computation of shadows~\cite{zhang2019shadowgan, liu2020arshadowgan}.
These techniques, however, are constrained to single timestep shadows, i.e., calculating scene shadows at a particular time.
While single-step computation might be useful for real-time shadow assessment at particular timesteps, analyses using the accumulation of shadows over multiple timesteps pave the way for environmental impact studies~\cite{tan2021urban} and city-scale urban planning~\cite{hodyl}.
Deep Umbra's focus is therefore on performant \emph{accumulation} of shadows over time, which requires innovative approaches to handle not only the spatial aspect (i.e., where casters of shadows are located), but also the temporal aspect (i.e., how shadows are moving over time) of the task.

In recent years, several open and proprietary solutions have emerged proposing analyses of the interplay between built environment and shadows, including online services~\cite{shadeMap,jveuxDuSoleil} and GIS tools~\cite{arcgis,qgis,miranda2018shadow}.
JveuxDuSoleil~\cite{jveuxDuSoleil} is a web-based tool for single timestep shadows.
ShadeMap is a closed source web-based tool that allows for the accumulation of shadows, though the on-the-fly computations are constrained to areas currently being rendered on screen, limiting comparisons across neighborhoods or cities.
Both QGIS and ArcGIS are fully-fledged GIS tools, but suffer from poor performance when accumulating shadows.
Google has recently introduced a paid API~\cite{solarApiGoogle} that provides roof's solar potential for a number of cities.
Our previous tool, Shadow Profiler~\cite{miranda2018shadow}, while providing significant gains when compared with previous efforts, still suffer from high latency.
Deep Umbra, on the other hand, offers performance gains and generalizability that allows for the accumulation of shadows over multiple cities, going beyond the scale tackled by previous approaches.

\highlight{Deep Umbra also adds to the growing number of open-source and low-cost solutions for urban planning.
These solutions include gamification approaches~\cite{angelidou2019social, kavouras2023low} and visual analytics toolkits~\cite{moreira2023urban, yap2023urbanity}.
Recently, Yap et al.~\cite{yap2022free} presented an extensive survey of open-source tools that support different stages of the urban planning process.}

\paragraph{GAN-assisted urban analytics}
The increase in urban data, combined with recent advancements in generative adversarial networks~\cite{10.1145/3394486.3403127,isola2017image}, has led to a number of works in the context of urban analytics~\cite{zheng2016visual}.
Wang et al.~\cite{10.1145/3397536.3422268} proposed a generative adversarial model for urban planning configuration, framing the problem as a task of learning land-use configurations given surrounding spatial contexts. 
Mokhtar et al.~\cite{mokhtar2020conditional} developed a conditional GAN model that is capable of generating pedestrian wind flow approximations for a variety of metropolitan topologies.
Zhang et al.~\cite{zhang2019trafficgan}, Zhang et al.~\cite{zhang2020cgail}, and Yuan et al.~\cite{yuan2022stgan} use GANs for traffic tasks.
Recently, Wu et al.~\cite{wu2022generative} reviewed applications of GANs in various urban tasks.
Our work is motivated by the growth of works using GANs to tackle urban-specific problems.
Deep Umbra is the first work that uses a generative adversarial approach to model shadow accumulation leveraging building height information and transferring knowledge across cities with varying urban morphologies.

\begin{figure*}[t]
  \centering
  \includegraphics[width=1\linewidth]{figs/overview.pdf}
  \caption{Overview the different components in the \name framework. Step 1, left: The preprocessing component is responsible for extracting building height information from OpenStreetMap~\cite{OpenStreetMap} (a) and generate $512\times 512$ tiles with this information (b). Step 2, right: The preprocessed data is then used to train a conditional generative adversarial network to map height $\rightarrow$ accumulated shadow, also taking into account season of the year and latitude of the tile (c). Cities used in the training set are shown as \protect\scalerel*{\includegraphics{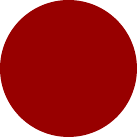}}{B} on the map, and cities used for testing are shown as \protect\scalerel*{\includegraphics{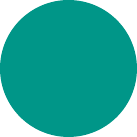}}{B}; \textbf{gt} refers to ground truth data and \textbf{ge} GAN generated data.}
  \label{fig:overview}
\end{figure*}

\section{Background
\label{sec:background}}



The study of shadow and sunlight access has been a core consideration in several domains, from architecture, urban planning, and civil engineering, to occupational therapy and environmental sciences.
A standard approach in these domains is to perform \emph{single-step shadow analysis}. In other words, only compute the shadow cast on public spaces for a very limited number of timesteps -- oftentimes only for a single timestep.
As we have highlighted in our previous work~\cite{miranda2018shadow}, there is a need by urban experts for a more comprehensive approach, moving beyond single-step to multi-step (i.e., accumulated) shadow analysis. 
For example, when comparing the sunlight access in two regions, rather than comparing across individual timestamps (e.g., shadows at every hour), an urban expert can simply compare the accumulated shadow in these two regions.
Data from the accumulation of shadows can inform different urban analyses, such as (1) urban heat island effect~\cite{tan2021urban}, (2) city planning (e.g., where to locate new parks)~\cite{hodyl}, and (3) public policy (e.g., restrictions on building heights)~\cite{central_park}.

The aforementioned concept is illustrated in Figure~\ref{fig:example}. Shadows considering a single timestep (left) offer a limited and superficial glimpse at the condition of sunlight access and shadow in public spaces. However, by accumulating shadows at multiple timesteps (right) we have a much more comprehensive understanding of the state of shadows and sunlight access in public spaces. 
The example shows the accumulation of shadows for 360 minutes, between 10~AM and 4~PM.  
Each pixel value is given by the sum of the shadows in each minute inside a time range:
\begin{align*}
A^{s,e}_{i,j} = \sum^{e}_{h=s}Shadow_{i,j}(h)\nonumber
\end{align*}

\noindent where $(i,j)$ is a point in the 2D plane, $s$ and $e$ are the start and end accumulation times, and $Shadow$ is a function that returns whether the point is under shadow or not at a particular timestep $h$.


%
In our example, the final accumulated value of shadows are then mapped to a colorscale, with darker shades of red indicating a larger amount of shadow coverage in the considered period.
Even though our proposal can be applied in the computation of arbitrary time ranges, in this paper we focus on three specific days: (1) summer solstice on June 21, (2) March equinox on March 20, (3) winter solstice on December 21. 
Two important considerations: it is enough to only consider a single equinox for both fall and spring and a summer solstice in the Northern Hemisphere is equivalent to a winter solstice in the Southern Hemisphere.
For each one of these days, we will consider accumulations one hour after sunset and one hour before sundown, totaling 720 minutes for (1), 540 for (2) and 360 for (3).

This comprehensive approach was also  featured in a \emph{New~York~Times} article~\cite{bui2016mapping}, stressing the need for the generation of this type of data not only for domain-specific purposes, but also for the engagement of a broader audience in this pressing issue.
In that light, three key aspects of our research are the ability to (1) leverage crowdsourced data that is globally available, (2) adhere to open and reproducible standards, and (3) ensure the transferability of our outcomes to cities in different countries and with varying built environment characteristics.

\section{Methodology}
\label{sec:methodology}

In this section, we describe the main components of our \name framework. At its core, \name consists of two components: (1) a pre-processing component that leverages crowdsourced information (e.g., OpenStreetMap~\cite{OpenStreetMap}) to generate tiles with building height information; (2) a conditional GAN model to infer the accumulated shadow values over different geographical regions and seasons. 

\subsection{Framework}
The overall framework of \name is shown in Figure~\ref{fig:overview}. 
%
%
In Step 1, we extract building height information from datasets that describe the urban morphology in a city. In this paper, we leveraged OpenStreetMap (OSM) data, but our approach is agnostic to the source, also being compatible with point cloud data and already-existing digital elevation models.
The geometry description is used to generate $256\times256$ image tiles, with each pixel containing the average building height at that particular location.
\highlight{To account for the fact that buildings from a tile can also cast shadows on neighboring tiles, we adopt a padding strategy. That is, for each $256\times256$ tile with height information (i.e., $b$), our framework pads it with neighboring height tiles, generating a $512\times512$ height tile that covers the central tile plus parts of its nine adjacent tiles. Figure~\ref{fig:padding} details this strategy.}

In Step 2, we train a model that is able to infer accumulated shadow information given the tiles from the previous step.
In this step, we leverage the $512\times512$ padded tiles.
In essence, our objective is to learn a function $\mathcal{G}: B \times L \times T \rightarrow A$, where $B$ is the space containing building information (i.e., height), $L$ represents the spatial information characteristic of a tile (i.e., latitude of a tile), and $T$ is the set of seasons. $A$ will then be the space of completed accumulated shadows. 
\name requires paired input and ground truth images for training.
As such, underlying our efforts is the understanding that \emph{data generation}, in our case, is straightforward yet computationally expensive.
Given a pre-processed tile (from Step 1), we can leverage off-the-shelf ray-tracing architectures to generate training data. In our case, we make use of \emph{Shadow Accrual Maps}~\cite{miranda2018shadow}, our previously introduced shadow accumulation algorithm.
Next, we describe both the model and architecture used to implement \name.

\subsection{Conditional GAN model}
\begin{figure}[t]
  \centering
  \includegraphics[width=1\linewidth]{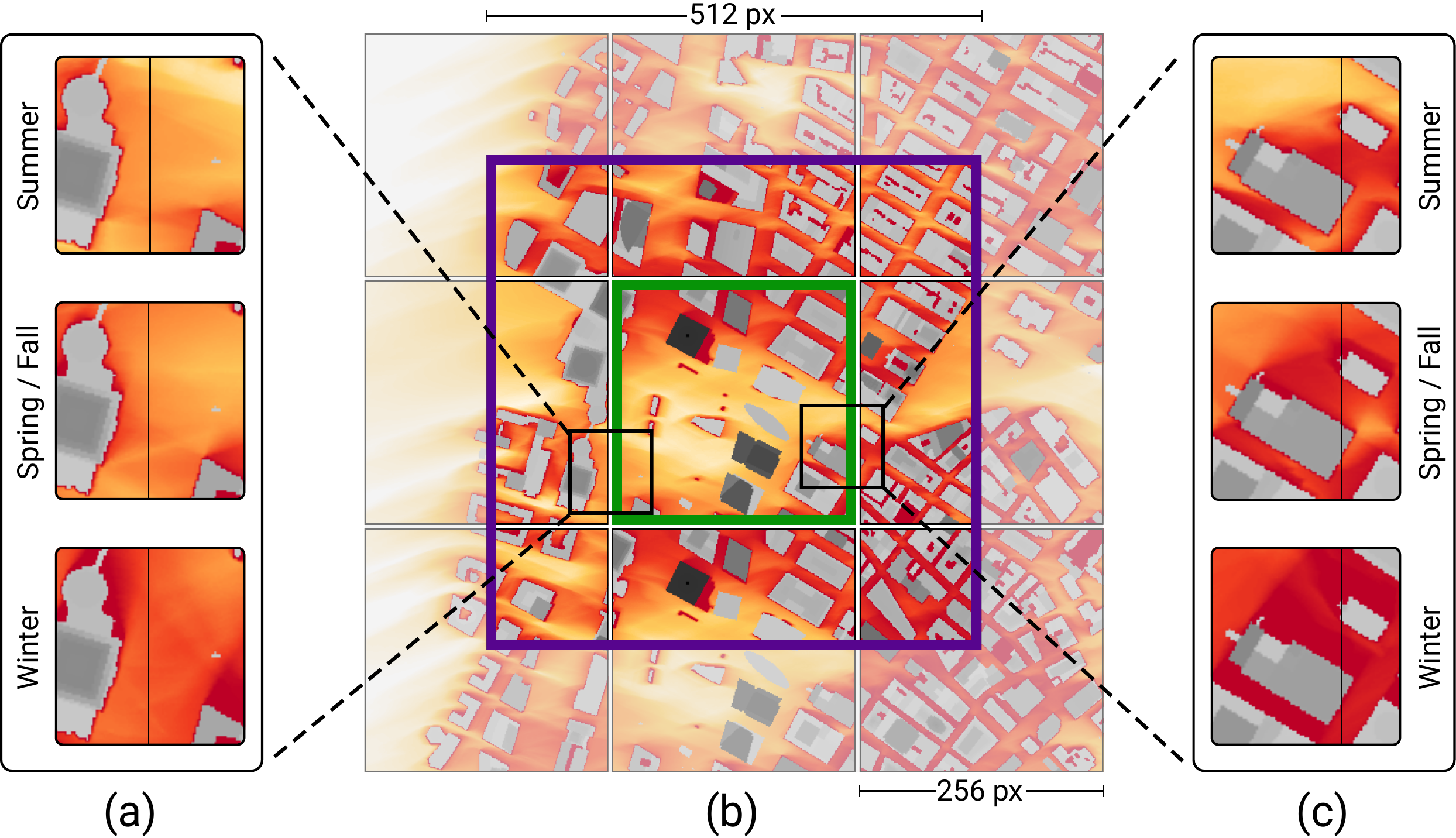}
  \caption{
  \highlight{Example where shadows are cast across tiles. In (b), the two small squares highlight areas where shadows from neighboring buildings are cast onto the center tile. This is further highlighted for different seasons in (a) and (c), with vertical black lines indicating the boundaries between tiles. To account for accumulated shadows from neighboring tiles, during training, a $256\times256$ height tile is padded with parts of adjacent tiles, resulting in a $512\times512$ tile (larger purple square in (b)). Model results are cropped back to the original size of $256\times256$ (smaller green square in (b)).}
  }
  \label{fig:padding}
\end{figure}

In order to predict the shadow tiles, we leverage the conditional GAN architecture~\cite{mirza2014conditional} to synthesize tiles with accumulated shadow. A generator in a conditional GAN learns a mapping from input data $x$ and random noise $z$ to an output image $a$, i.e., $G: {x,z} \rightarrow y$~\cite{isola2017image}. 
A discriminator $D$ will be trained to distinguish between synthesized and real tiles.
More specifically, in our case, the generator $G$ will consider: (1) height tiles $b \in B$ padded with neighboring tiles (as previously mentioned and highlighted in Figure~\ref{fig:padding}); each $512\times512$ padded tile $b$ represents the building morphology of the location covered by the tile. (2) spatial information $l \in L$ of the tiles, with the associated latitudes of the tile. (3) temporal information $t \in T$, with one of the three considered accumulation periods (summer solstice, March equinox, winter solstice).
With (1), \name takes into account building morphology, while with (2) and (3) it takes into account the apparent movement of the sun across the sky (i.e, the sun path) at different latitudes and seasons.
$G$ will then generate shadow tiles $\hat{a}$ seeking to mimic the ground truth from that region ($a \in A$).
In short, we have:
\begin{align*}
    x &= (b,l,t)\\
    \hat{a} &= G(x)
\end{align*}

The discriminator $D$ is trained to classify real and synthetic tiles when compared to the ground truth, whereas the generator $G$ tries to fool the discriminator by generating images as close to the ground truth as possible. To increase diversity in the generation task, $G$ uses dropouts for implicit noise~\cite{krizhevsky2012imagenet}.
This contest between $D$ and $G$ will be formulated through an objective that contains a reconstruction loss $\mathcal{L}_{r}$ and an adversarial loss $\mathcal{L}_{adv}$, detailed next.

\paragraph{Reconstruction loss} 
A reconstruction loss will try to reduce the pixel-wise error between ground truth and the generated tile. Traditional approaches usually rely on $L_1$ or $L_2$ distances as the reconstruction loss~\cite{pathak2016context, isola2017image}. 
In our work, we experimented with both and found out that $L_1$ performs better than $L_2$.
However, both failed to capture the sharpness and detail of long shadows and instead produced blurry results, especially for the longer winter shadows at higher latitudes.
To minimize these issues, we augmented the usual $L_1$ loss with a structural similarity loss function (SSIM loss) to better capture details in spatially adjacent pixels~\cite{wang2004image}.
In addition, we also examined an edge detection-based Sobel loss function to preserve the sharpness in longer shadows~\cite{paul2022edge}.
Following previous works that combine multiple loss functions~\cite{liu2020arshadowgan, paul2022edge, wang2022esa, wang2018high}, we performed a set of experiments and found out that a combination of $L_1$, SSIM, and Sobel losses attains the best results at generating accumulated shadow information. We discuss these in more detail as part of the ablation study in Section \ref{sec:experiments}. In short, $\mathcal{L}_{r}$ is given by:

\begin{align*}
    \mathcal{L}_{r}(G) &= \mathcal{L}_{1}(G)+\mathcal{L}_{SSIM}(G)+\mathcal{L}_{Sobel}(G)\\
    \mathcal{L}_{1}(G) &= \|a-\hat{a}\|_1\\
    \mathcal{L}_{SSIM}(G) &=1-SSIM(a,\hat{a})\\
    \mathcal{L}_{Sobel}(G) &=(Sobel(a)-Sobel(\hat{a}))^2
\end{align*}

\paragraph{Adversarial loss}
The competition between discriminator $D$ and generator $G$ can be formulated by the adversarial loss function as follows:

\begin{align*}
    \mathcal{L}_{adv}(G,D) = \mathbb{E}_{x,a}[logD(x,a)] + \mathbb{E}_{x}[log(1 - D(x, \hat{a}))]\nonumber
\end{align*}

During training, the generator $G$ tries to minimize ${L}_{adv}$, whereas the goal of discriminator $D$ is to maximize it.
This means that the generated shadow tile $\hat{a}$ should be as close to the ground truth $a$ as possible.
The discriminator $D$ takes as input either the generated shadow tiles $\hat{a}$ or the ground truth tiles $a$ to determine whether $\hat{a}$ is real or not.

\paragraph{Full objective}
The overall objective function of our conditional GAN is composed of the adversarial loss $\mathcal{L}_{adv}$ and reconstruction loss $\mathcal{L}_{r}$ and can be expressed as following:

\begin{align*}
    G^{*} = \mathrm{arg}\  \underset{G}{\mathrm{min}}\  \underset{D}{\mathrm{max}}\ \mathcal{L}_{adv}(G,D) + \lambda \mathcal{L}_{r}(G)\nonumber
\end{align*}

\noindent where $\lambda$ controls the relative importance of reconstruction loss compared to the adversarial loss. We use $\lambda=100$.


\subsection{Architecture \& implementation}
To satisfy our requirements and similar to previous conditional GAN approaches, we leverage PatchGAN~\cite{isola2017image} as the discriminator $D$.
For the generator, we tested two architectures: U-Net~\cite{ronneberger2015u} and ResNet~\cite{jian2016deep}, ultimately choosing ResNet as our generator $G$.
%
%
%
It is important to stress that conditional GANs are difficult to train and sensitive to hyperparameters.
Given that \name is the first attempt to tackle the problem of generating accumulated shadow maps with spatial and temporal information, we chose to leverage well-known architectures (PatchGAN + ResNet), which in turn allowed us to rely on already established hyperparameters and training procedures.
In addition, Section~\ref{sec:experiments} presents a detailed ablation study. In it, we discuss the selection of the architecture and the loss function used in our final model.

\name's data ($b,l,t,a,\hat{a}$) follows the tile map format to facilitate implementation in the web environment (e.g., web maps). As previously mentioned, the input and ground truth data are padded tiles with size $512\times512$ to account for shadows from neighboring tiles. 
The entire training process (Figure~\ref{fig:overview} (right)) uses these $512\times512$ tiles.
%
\highlight{To adhere to the tile map size of $256\times256$, we make use of a cropping operator $C$ that removes the outer tiles from the model result ($\hat{a}$), leaving only the central $256\times256$ tile ($C(\hat{a})$), as depicted in Figure~\ref{fig:padding}.}


\highlight{
We use ReLU activation in the generator and Leaky ReLU in the discriminator layers, also incorporating stride in the discriminator for downsampling and fractional stride in the generator for upsampling. Moreover, we apply batch normalization in both the generator and discriminator.
Adam optimizer is used with a learning rate of 0.0002 and a momentum of 0.5.
}

%
%


\section{Data Description}
\label{sec:data}
We now describe the pipeline to extract height information from OSM data, and the process to compute the ground truth data used to train \name.

\begin{figure}[t]
  \centering
  \includegraphics[width=1\linewidth]{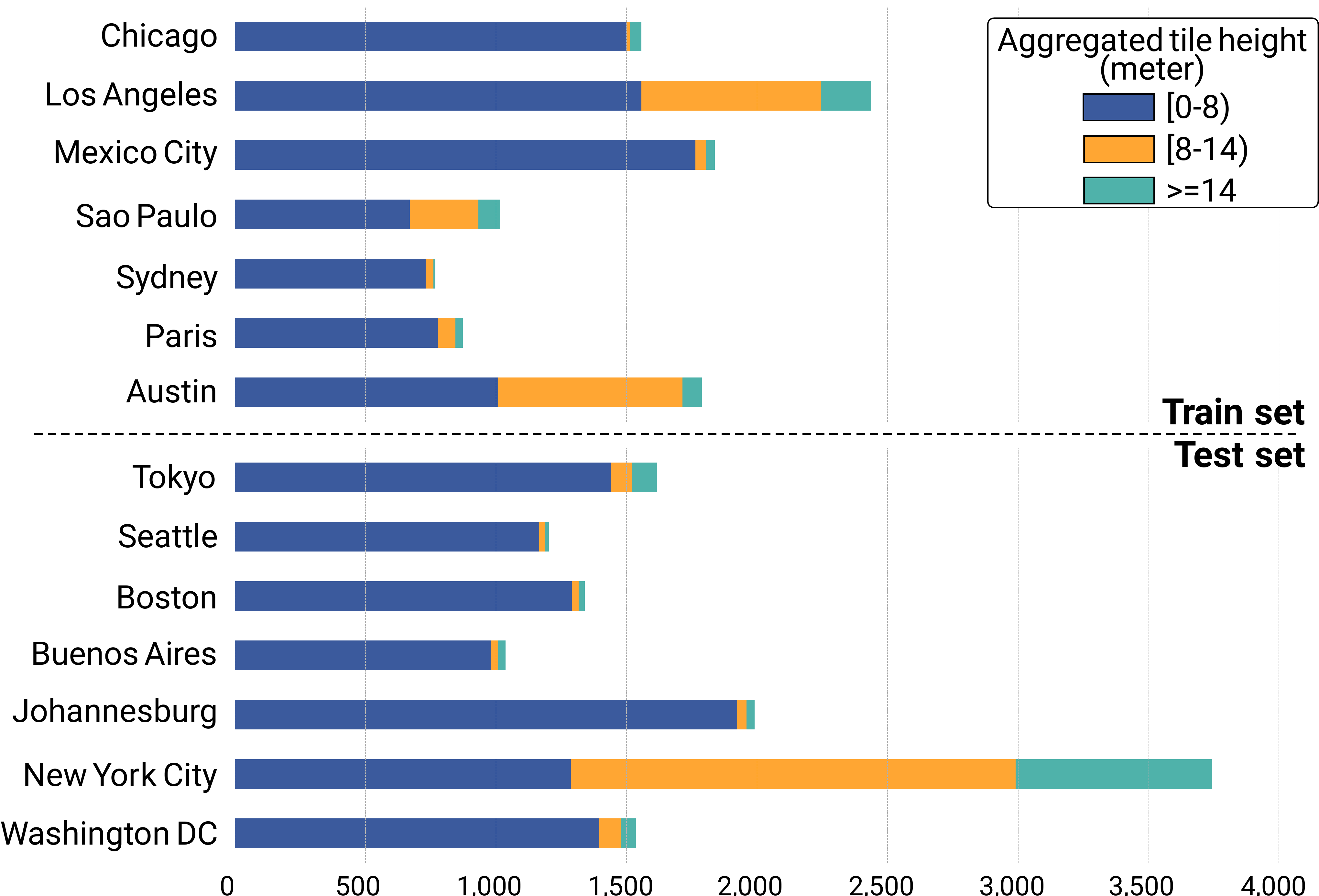}
  \caption{Distribution of average tile heights for the train and test sets.}
  \label{fig:building_height}
\end{figure}

\paragraph{Building height data}
%
%
For each city of interest, we first determine the bounding box that covers the entire spatial extension of the urban area and use this information to query OSM for building geometry information.
After that, we generate tiles (with size $256 \times 256$ pixels) covering the entire region and storing the mean building height in each pixel -- we normalize this data considering a maximum building height of 500 meters.
In this paper, we focus on tile maps using zoom level 16. This zoom level gives us a good tile resolution of $\sim$2.3 meters per pixel.
Although we rely on OSM for height information, we realize that crowdsourced initiatives have inherent limitations in terms of spatial coverage. However, we note that our approach can leverage any data source that can generate reliable height information -- such as point clouds and high-resolution digital elevation models.

\paragraph{Ground truth shadow data}
To create the ground truth data to train our model, we use our previous work~\cite{miranda2018shadow}, a GPU-based ray-tracing technique that is optimized for accumulating shadow values.
Due to its interpolation approach, it has considerable performance gains even when compared with industry standards, such as Nvidia's OptiX.

%


\section{Experiments \& Results}
\label{sec:experiments}

To evaluate our framework, we performed experiments considering a large set of cities, taking into account different continents and building morphologies.
Our evaluation can be divided into four parts.
First, we performed an ablation study to properly evaluate and choose the best configurations for \name.
Second, we quantitatively evaluated \highlight{the best configuration from the ablation study} on a number of cities across different continents, including cities outside the initial training set.
Third, we qualitatively evaluated \name, highlighting both general cases as well as large error cases.
Finally, we compared \name against existing tools and techniques, including commercial and open-source ones.
We report root mean square error (RMSE), mean absolute error (MAE), mean squared error (MSE), and structural similarity (SSIM), considering ground truth tiles as a reference to the generated shadow tiles.

To ensure generalizability, \highlight{our quantitative experiments follow two approaches: first, tiles from the same set of cities are used for the training and test sets (i.e., \emph{within cities evaluation}), and second, tiles from two disjoint sets of cities are used for the test and training sets (i.e., \emph{across cities evaluation}).}
The cities were carefully picked to ensure that they had buildings with different heights and morphologies. 
Figure~\ref{fig:building_height} shows the distribution of mean building height per tile for the train and test cities used in the experiments. 
To account for the scant number of tiles with large mean building height ($>=14$ meters), we sample the training data with 50\% tiles with mean building height below or equal to the 50th percentile and the rest above the 50th percentile.


%
%
%
%
%
%

We implemented our model using TensorFlow. The experiments were executed on a desktop computer with an Intel i7-13700KF CPU, 32~GB of RAM, and an NVIDIA GeForce 3080 with 10~GB of RAM.

\begin{figure}[t]
  \centering
  \includegraphics[width=1\linewidth]{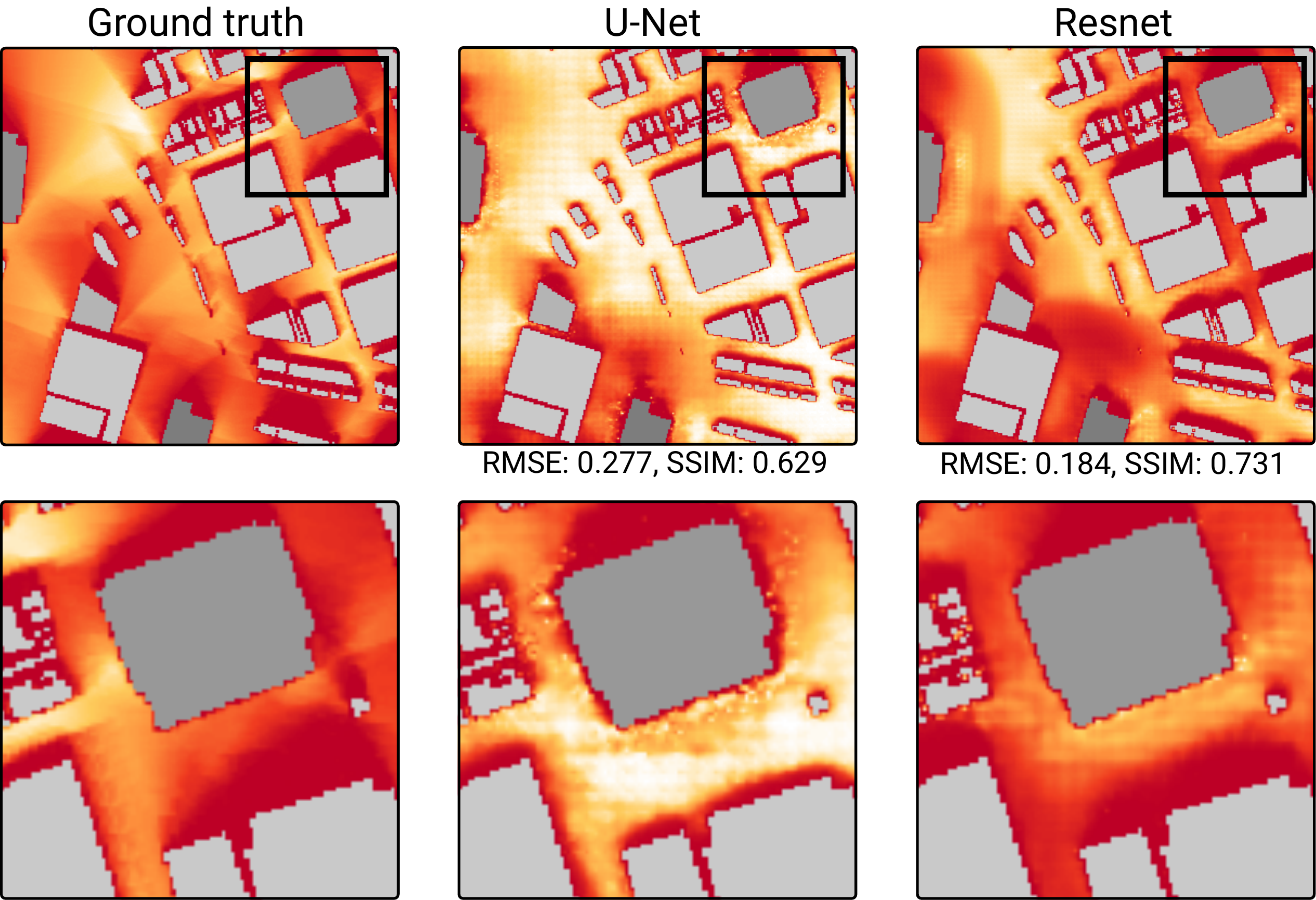}
  \caption{Results of ablation study using U-Net and ResNet generator architecture and $L_1$ loss function. RMSE and SSIM scores with respect to ground truth are highlighted below each image (for SSIM, higher values are better). Note that ResNet architecture produces significantly better results, especially for longer shadows that are distant from the building. Top: Highlights with areas with longer shadows. Bottom: Zoomed areas.}
  \label{fig:ablation_gen}
\end{figure}

\begin{table}[t]

\ttabbox{%
\caption{Results of ablation experiments with different generator architectures. For SSIM, higher values are better.}
\centering
\begin{tabular}{c|c|c|c|c}
Generator & RMSE & MAE & MSE & SSIM \\
\hline
U-Net & 0.0706 & 0.0308 & 0.0073 & 0.8619\\
\textbf{ResNet} & \textbf{0.0693} & \textbf{0.0303} & \textbf{0.0071} & \textbf{0.8629}\\
\hline
\end{tabular}
\label{table:ablation_gen}
}%

\ttabbox{%
\caption{Results of ablation experiments with different loss functions and ResNet generator. For SSIM, higher values are better.}
\centering
\begin{tabular}{c|c|c|c|c}
Loss function & RMSE & MAE & MSE & SSIM \\
\hline
$L_1$ & 0.0693 & 0.0303 & 0.0071 & 0.8629\\
$L_2$ & 0.0703 & 0.0328 & 0.0072 & 0.8446\\
SSIM + $L_1$ & 0.0667 & 0.0292 & 0.0067 & 0.8718\\
SSIM + Sobel & 0.0653 & 0.0296 & 0.0065 & 0.8657\\
\textbf{SSIM + Sobel + $L_1$} & \textbf{0.0643} & \textbf{0.0289} & \textbf{0.0063} & \textbf{0.8724}\\
\hline
\end{tabular}
\label{table:ablation_loss_func}
}%

\end{table}%

\begin{figure*}[t]
  \centering
  \includegraphics[width=1\linewidth]{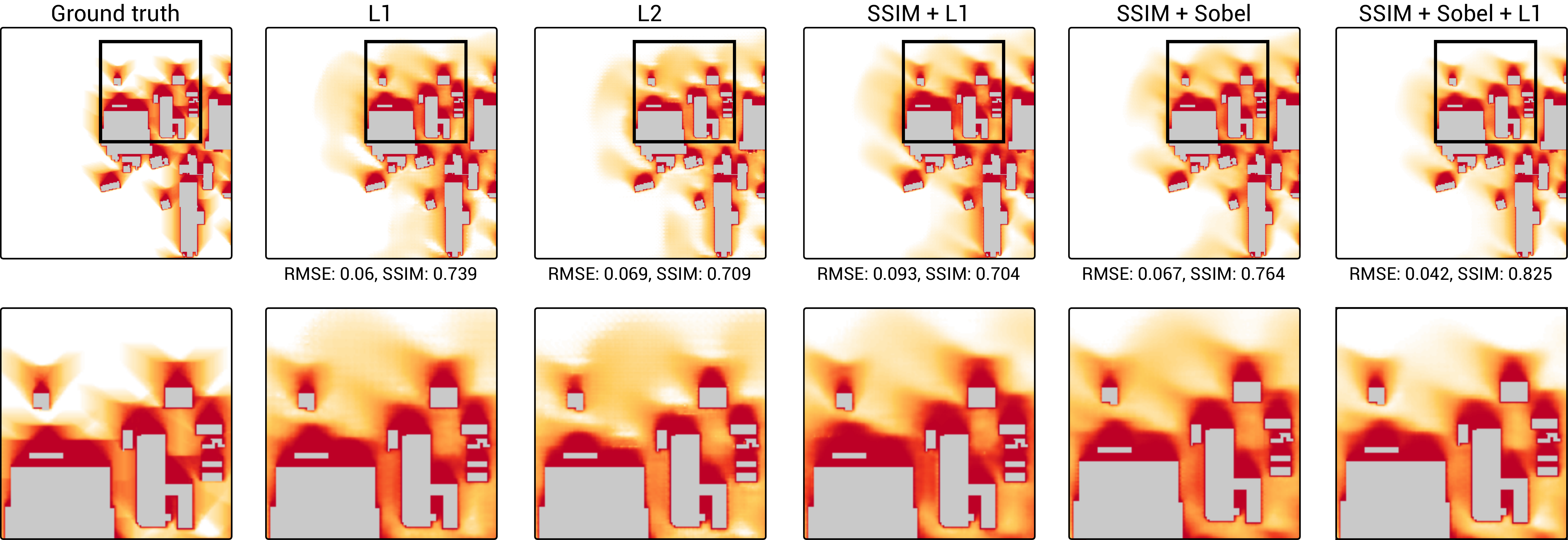}
  \caption{Results of ablation study testing different loss functions with the ResNet architecture. RMSE and SSIM scores with respect to the ground truth are highlighted below each image (for SSIM, higher values are better). SSIM and Sobel losses combined with $L_1$ distance generate more precise accumulated shadows, whereas combinations of other loss functions are blurrier. We note that complex shapes (such as the \emph{Y}-like pattern due to the sun's winter path) are more accurately represented with the aforementioned combination. Top: Highlights with areas with longer shadows. Bottom: Zoomed areas.}
  \label{fig:ablation_loss_funcs}
\end{figure*}

\subsection{Ablation study}

\highlight{
As a baseline, we considered a simple model with a U-Net generator and $L_1$ loss function that solely utilizes building height ($b$) as input and does not consider latitude ($l$) and season ($t$). The RMSE, MAE, MSE, and SSIM scores for the baseline model are 0.1171, 0.0556, 0.0208, and 0.7729, respectively.}
\highlight{The next configurations incorporate all three parameters ($b, l, t$) as the input for the model.}
\highlight{Our ablation study uses the \emph{within cities evaluation} setting (see Section~\ref{sec:quantitative}).}


Although U-Net is the most commonly used architecture for image-to-image translation~\cite{isola2017image}, Hu et al.~\cite{hu2018image} have shown that residual-based generators outperform U-Net in certain tasks. 
Thus, to verify \name's effectiveness, we tested two generator architectures: U-Net~\cite{ronneberger2015u} and ResNet~\cite{jian2016deep}. In both cases, we added $L_1$ distance along with the adversarial loss to generate outputs closer to the ground truth. The results, summarized in Table \ref{table:ablation_gen}, reveal \highlight{that both configurations outperformed the baseline model and that the ResNet architecture had better results than U-Net}.
We further visually inspected the results and found that, while both architectures perform fairly well for shadows closer to the buildings, ResNet performs significantly better in inferring longer shadows, especially for tiles from the winter season (see Figure \ref{fig:ablation_gen}).

Next, we tested alternative loss functions with the ResNet architecture to further evaluate the model. $L_1$ and $L_2$ distances are standard loss functions used in image-to-image translation tasks~\cite{pathak2016context, isola2017image}. 
In addition, we tested combinations of SSIM and Sobel losses to (1) preserve the structural integrity and (2) improve the sharpness of shadows, especially in long shadows. 
$L_1$ and $L_2$ distances only focus on error sensitivity, whereas SSIM loss, as an alternate measure, evaluates the structural, luminance, and contrast difference between ground truth and generated shadow images~\cite{wang2004image}. 
Furthermore, Sobel loss tends to reduce the inherent blurs in shadows~\cite{paul2022edge}. The results, summarized in Table \ref{table:ablation_loss_func}, reveal that a combination of SSIM, Sobel, and $L_1$ loss produces the best outputs. 
We also visually inspected the results and found out that by combining SSIM and Sobel with $L_1$ loss, the generated images most closely depicted the ground truth compared to $L_1$ loss being used alone (see Figure~\ref{fig:ablation_loss_funcs}).
For the next experiments, we used the ResNet model coupled with SSIM, Sobel, and $L_1$ losses.


\subsection{Quantitative evaluation}
\label{sec:quantitative}

To quantitatively assess \name's error, we performed two evaluations. In the first evaluation, we selected tiles from seven cities to create training and test sets. Then, in the second evaluation, we used the trained model in seven other cities (outside the training set).

\paragraph{Within cities evaluation}
In this evaluation, we selected seven cities from four different continents and with varying building morphologies: Los Angeles, Chicago, Austin, Mexico City, S\~{a}o Paulo, Paris, and Sydney (\protect\scalerel*{\includegraphics{figs/circle-red_svg-tex.pdf}}{B} circles in Figure~\ref{fig:overview}~(left)).
In total, we generated ground truth (i.e., ray-traced) data for 20,260 height tiles of these cities.
Then, we randomly selected 750 tiles from \emph{each} city in \emph{each} season (i.e., $750 \times 3$ tiles per city). We then used k-fold cross validation with $k=5$ for training and testing.
Table~\ref{table:within} shows the average RMSE, MAE, MSE, and SSIM across all 5 runs.
RMSE is fairly low at 0.0631, which in practical terms means that, when accumulating 360 minutes for the winter season, the error would be equal to approximately 23 minutes.

%
%

\begin{table}[t]
\ttabbox{%
\caption{Within cities analysis. For SSIM, higher values are better.}
\centering
\begin{tabular}{c|c|c|c|c}
& RMSE & MAE & MSE & SSIM\\
\hline
K-fold ($k=5$) & 0.0631 & 0.0276 & 0.0063 & 0.9000\\
\hline
\end{tabular}
\label{table:within}
}%

\ttabbox{%
\caption{Across cities analysis. For SSIM, higher values are better.}
\centering
\begin{tabular}{l|c|c|c|c}
Target city & RMSE & MAE & MSE & SSIM\\
\hline
Washington DC & 0.0786 & 0.0318 & 0.0091 & 0.8819\\
NYC & 0.0818 & 0.0374 & 0.0084 & 0.8792\\
Boston & 0.0743 & 0.0309 & 0.0081 & 0.8985\\
Seattle & 0.0597 & 0.0339 & 0.0043 & 0.8578\\
Johannesburg & 0.0281 & 0.0102 & 0.0014 &  0.8749\\
Buenos Aires & 0.0441 & 0.0146 & 0.0033 &  0.9160\\
Tokyo & 0.0837 & 0.0432 & 0.0093 & 0.7981\\
\hline
Average & 0.0643 & 0.0289 & 0.0063 & 0.8724\\
\hline
\end{tabular}
\label{table:across}
}

\ttabbox{%
\caption{Across cities performance analysis, considering street networks.}
\centering
\begin{tabular}{l|c|c|c|c}
Target city & RMSE & MAE & MSE & SSIM\\
\hline
Washington DC & 0.0262 & 0.0053 & 0.0013 & 0.9579\\
NYC & 0.0289 & 0.0067 &	0.0012 & 0.9593\\
Boston & 0.0263 & 0.0054 & 0.0013 & 0.9618\\
Seattle & 0.0249 & 0.0067 & 0.0009 & 0.9452\\
Johannesburg & 0.0095 & 0.0017 & 0.0003 & 0.9697\\
Buenos Aires & 0.0162 & 0.0027 & 0.0005 & 0.9739\\
Tokyo & 0.0447 & 0.0135 & 0.0029 & 0.9172\\
\hline
Average & 0.0252 & 0.0060 & 0.0012 & 0.9550\\
\hline                        
\end{tabular}
\label{table:street_metric}
}%

\end{table}

\paragraph{Across cities evaluation}
In this second evaluation, we selected seven different cities to assess the generalizability of the model trained in the \emph{within cities evaluation}.
The seven selected cities were Washington DC, NYC, Boston, Seattle, Johannesburg, Buenos Aires, and Tokyo.
In this evaluation, we also selected 750 tiles from each city in each season (i.e., 2,250 tiles per city).
Table~\ref{table:across} presents the RMSE, MAE, MSE, and SSIM in each one of the seven target cities -- again the model was not trained with data from these cities.
High-density cities such as Washington DC, NYC, and Tokyo present the highest RMSE ($\sim$0.08), and Tokyo, in particular, presents the lowest SSIM value.
Given the importance of street-level analyses for urban accessibility~\cite{hosseini_towards_2022,froehlich_future_2022,hosseini_mapping_2023}, we further tested \name only considering tile pixels that fall within streets (and not parks or other facilities). The results (Table~\ref{table:street_metric}) show that, when only considering streets, our results show even lower errors (mean RMSE of $\sim$0.02).

\begin{figure*}[t]
  \centering
  \includegraphics[width=1\linewidth]{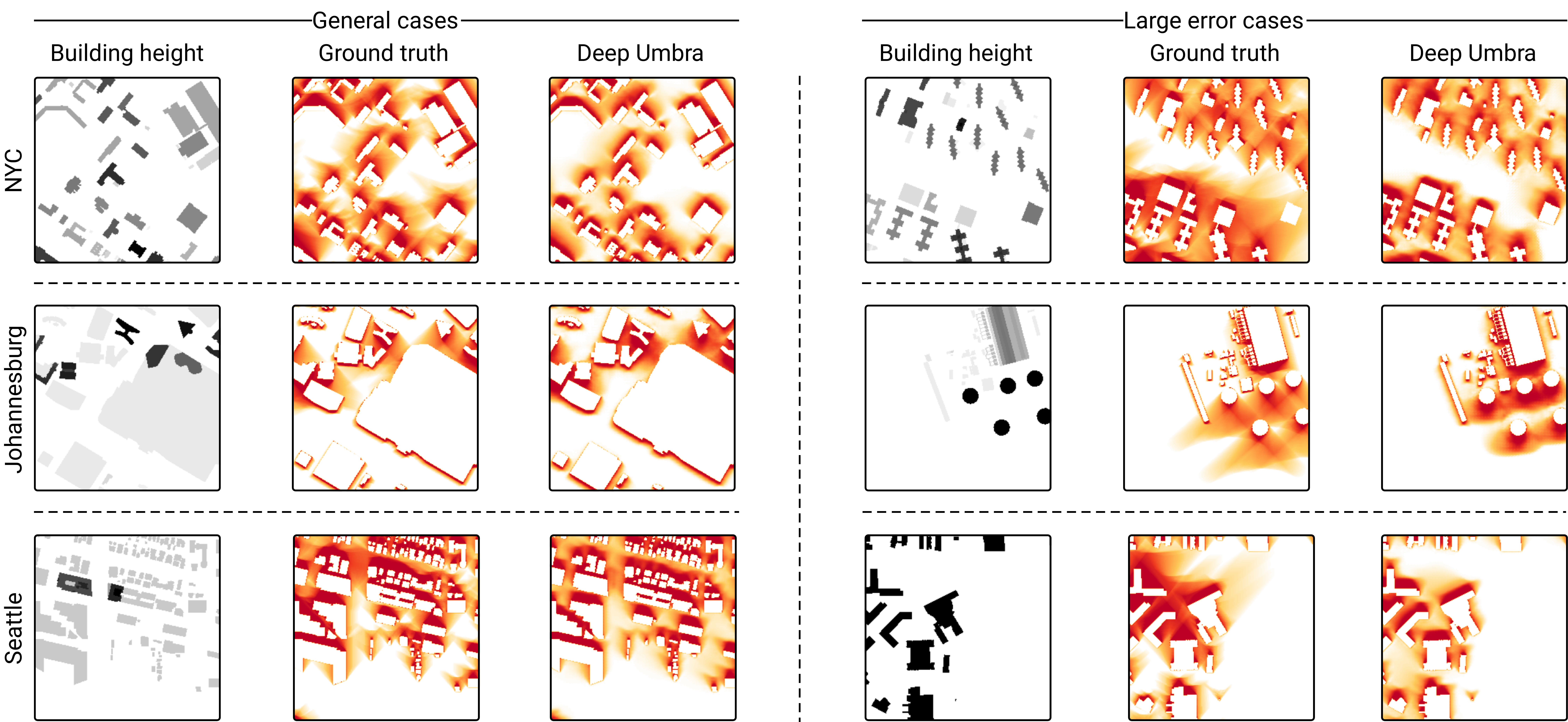}
  \caption{Results with tiles from NYC, Johannesburg, and Seattle. The examples come from different seasons of the year, but use the same colorscale where darker shades of red correspond to greater shadow coverage during the accumulation period. The left column shows general cases where the RMSE is within the interquartile range. The right column shows large error cases where the RMSE is larger than $1.5\times$ the interquartile range.}
  \label{fig:results}
\end{figure*}

\subsection{Qualitative evaluation}
We also highlight a set of qualitative results from our trained model using tiles from different cities. 
Figure~\ref{fig:results} shows relevant images for NYC, Johannesburg, and Seattle, including building height input, ground truth and \name results.
In these cases, it is possible to notice that the generated shadow values more closely match the ground truth when pixels are located closer to buildings (Johannesburg, left; Seattle, left). Yet, when moving away from them, many of the nuances and sharpness of the shades are lost (NYC, right; Johannesburg, right).
Figure~\ref{fig:results} also highlights large error cases where the RMSE is larger than $1.5\times$ the interquartile range. We note, however, that these cases represent less than 5\% of the total cases in our test set, underscoring how well our framework is able to model shadow accumulation.

\begin{table*}[t]
\caption{
Comparison between different shadow assessment tools and techniques. We classify as \emph{interactive} the solutions that compute results in less than $\sim$0.15 second, and \emph{real-time} the ones with results below $\sim$0.05 second.\emph{Open source} refers to software that is freely available for code inspection.}
\centering
\begin{tabular}{c|c|c|c|c|c}
Techniques \& tools & Performance & What-if support & Accumulation & Accuracy &  Open source\\
\hline
Shadow Accrual Maps~\cite{miranda2018shadow}  & Interactive & \textbf{Yes} & \textbf{Yes} & \textbf{Exact} & No\\
\hline
JveuxDuSoleil~\cite{jveuxDuSoleil} & \textbf{Real-time} & No & No & \textbf{Exact} & No\\
\hline
ShadeMap~\cite{shadeMap} & Interactive & \textbf{Yes} & \textbf{Yes} & \textbf{Exact} & No\\
\hline
QGIS~\cite{qgis} & Non-interactive & \textbf{Yes} & \textbf{Yes} & \textbf{Exact} & \textbf{Yes}\\
\hline
ArcGIS Pro~\cite{arcgis} & Non-interactive & \textbf{Yes} & \textbf{Yes} & \textbf{Exact} & No\\
\hline
Google Solar~\cite{solarApiGoogle} & Offline (pre-computed) & No & \textbf{Yes} & \textbf{Exact} & No\\
\hline
\textbf{Deep Umbra (this work)} & \textbf{Real-time} & \textbf{Yes} & \textbf{Yes} & $\sim$0.06 RMSE & \textbf{Yes}
\end{tabular}
\label{table:comparison}
\end{table*}

\subsection{Comparison with existing tools \& techniques}
We also compared \name with existing tools and techniques across five dimensions: accuracy, performance, support for what-if analyses, support for shadow accumulation, and whether the solution was open or not.
In collaboration with urban experts (co-authors of the paper), we selected five already-existing tools and techniques for comparison:
(1)~Shadow Accrual Maps~\cite{miranda2018shadow}, a GPU-based technique for the interactive accumulation of shadows at city scale and achieves a speedup of 10$\times$ when compared to Nvidia's OptiX.
(2)~JveuxDuSoil~\cite{jveuxDuSoleil}, a web-based service for the assessment of single timestep shadows.
(3)~ShadeMap~\cite{shadeMap}, a web-based service that supports the accumulation of shadows.
(4)~QGIS~\cite{qgis} and ArcGIS Pro~\cite{arcgis}, two popular GIS tools in urban planning and architecture that also support shadow accumulation.
(5)~Google~Solar~\cite{solarApiGoogle}, an API that allows for the querying of hourly shade information.
Table~\ref{table:comparison} summarizes the comparison.

Regarding time performance, \name and JveuxDuSolei are the only ones that achieve real-time rates, i.e., the computations are performed in less than $\sim$0.05 second. Lower times are particularly important for two reasons: First, to drive visual analytics systems, as increased latency reduces users' rate of observations, generalizations and generation of hypotheses~\cite{liu_effects_2014}. 
Second, to enable the fast computation of accumulated shadows for larger geographical regions. For example, to compute the shadow ground truth data for the 7,000 tiles in NYC, Shadow Accrual Maps took more than 12 minutes for a single season. Such a situation is even more impractical when considering GIS tools, such as QGIS. Using the Urban Multi-scale Environmental Predictor (UMEP) extension, it took 3 hours to accumulate shadows for a single season. \name, on the other hand, was able to compute the accumulated shadows for the entire city within 105 seconds, a speed up of 6$\times$ compared to Shadow Accrual Maps, and 102$\times$ compared to QGIS.

Support for what-if analyses was also considered in our comparison, i.e., the ability to replace buildings and re-compute analyses to assess shadow impact of proposed developments.
JveuxDuSoil and Google Solar do not support such analyses.
JveuxDuSoil also does not support shadow accumulation.
Of all techniques and solutions, \name is the only one to support real-time accumulation of shadows for what-if analyses.
Considering accuracy, previous tools support the exact computation of shadows (considering user-specified temporal and spatial resolutions). While not being exact, \name provides an alternative with fairly low error and significant time performance gains.
\name, together with QGIS, is the only solution that is open source and freely available for code inspection and modification.

\section{\dataset}
\label{sec:dataset}

As part of our work, we have also made available a large collection of accumulated shadow tiles for over 100 cities (in 6 continents), computed using \name.
The complete dataset contains 999,807 map tiles, totaling over 16~GB.
The selected cities cover six continents and have different building morphologies and urban street networks.
We hope that the data will provide new opportunities for urban experts to perform not only fine-grained analyses (i.e., \emph{what is the best location for a certain facility?}) but also large-scale comparisons and exploratory analyses across neighborhoods or cities (e.g., \emph{what is the park with least shadow amount?} or \emph{what is the city with least shadow amount?}).
Our objectives are twofold: (1) facilitate research on the topic of sunlight and shadow quantification by providing a dataset that can be compared to and augmented; and (2) enable research across different domains, particularly urban planning, and public health.
We have also made available a web-based interface that can be used to effortlessly browse and visualize the \dataset. 
\highlight{
The dataset currently contains accumulated shadow information, but the accumulated sunlight can be obtained by simply computing the complement of the accumulated shadow value.
}
\highlight{The dataset and web viewer} can be accessed at \href{http://urbantk.org/shadows}{urbantk.org/shadows}.


\begin{figure*}[t]
  \centering
  \includegraphics[width=1\linewidth]{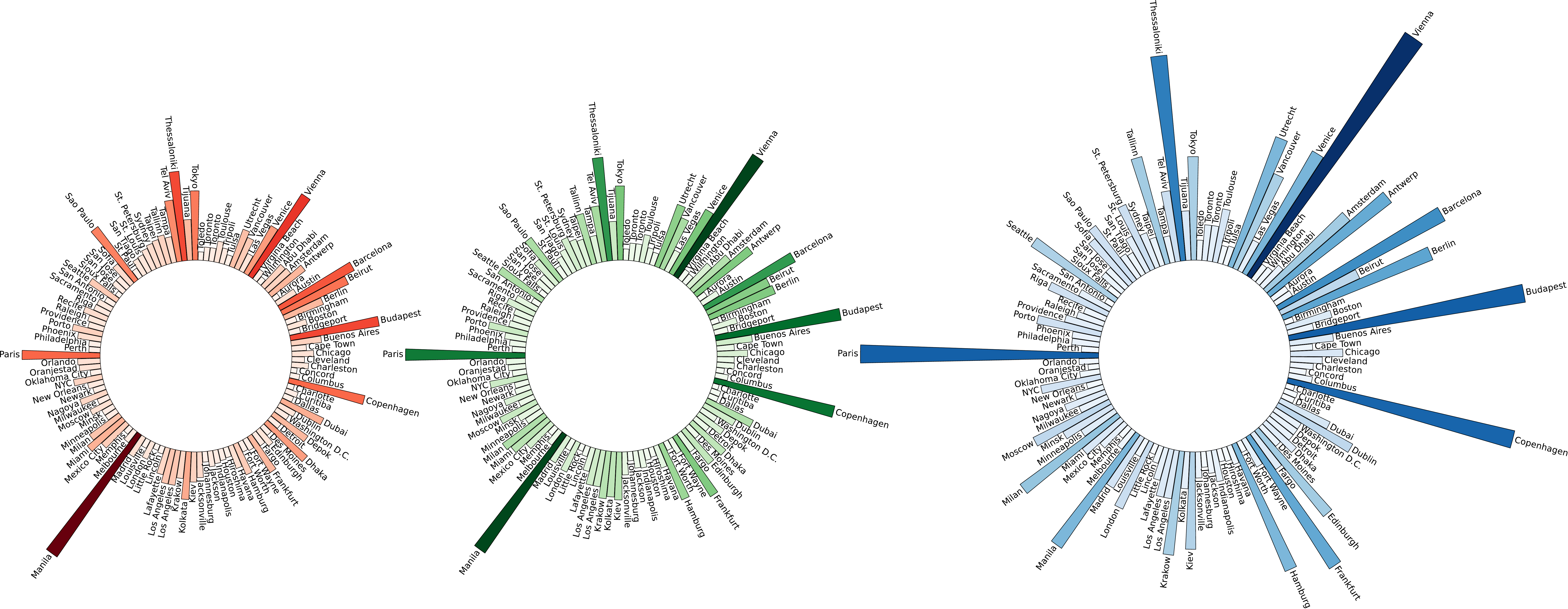}
  \caption{Comparison of mean shadow accumulation scores across parks in 110 cities during the summer (left), spring (middle), and winter (right).}
  \vspace{-0.25cm}
  \label{fig:allcities}
\end{figure*}

\section{Case Studies}

\label{sec:cases}

In this section, we demonstrate the application of our proposed framework through a set of case studies performed in collaboration with urban experts, both co-authors of this paper.
First, to highlight the scale and opportunities brought forward by \dataset, Figure~\ref{fig:allcities} visualizes the mean shadow accumulation for parks in 110 cities.
As expected, cities near the equator present similar values across seasons. Cities in high latitudes, however, have vastly different values between summer and winter. Specifically, Paris exhibited relatively high shadow values during winter.
The first case study then analyzes the impact of building density on winter sunlight access in several parks in the Vaugirard neighborhood in Paris. 

\begin{figure}[t]
  \centering
  \includegraphics[width=1\linewidth]{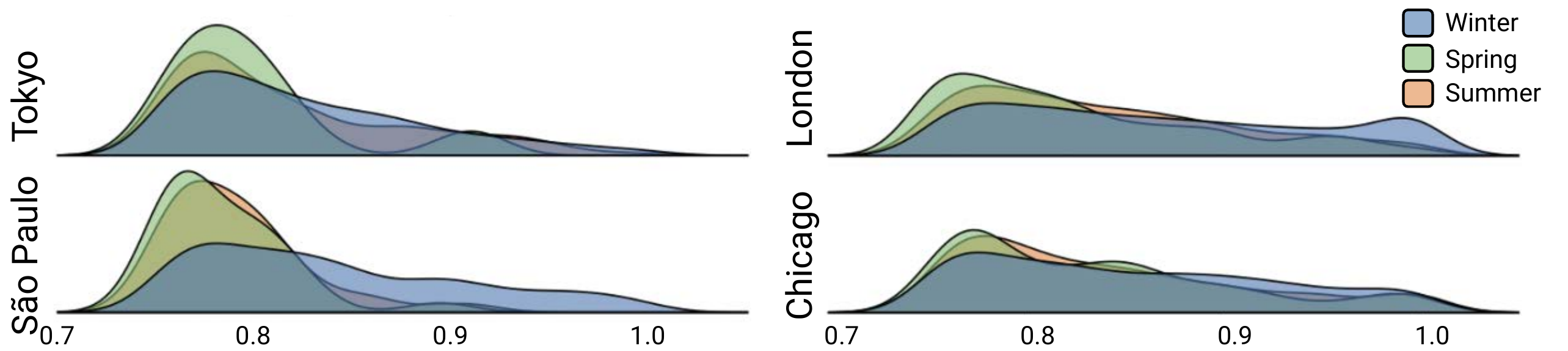}
  \caption{Distribution of shadow accumulation in overshadowed streets.}
  \label{fig:street}
\end{figure}

We further illustrate the utility of our work by measuring shadow accumulation on street networks, commonly utilized in urban accessibility studies~\cite{hosseini_towards_2022,froehlich_future_2022,hosseini_mapping_2023}. Figure~\ref{fig:street} visualizes the distributions of overshadowed streets (i.e., streets with mean shadow accumulation value above 75\%) over three seasons for four cities. We can see that for these cities, the shadow distribution in the winter is more skewed towards higher values than summer/spring, partially due to the sun's lower elevation during the season. Our second case study is then based on an in-depth analysis of the shadow accumulation on streets aggregated by the nearby land use type (e.g., residential, commercial, industrial) in two of the largest cities in the US: Chicago and NYC.

\begin{table}[t]
\caption{Percentage of parks that have high sunlight access (mean shadow accumulation value below 25\%), moderate sunlight access (between 25\% and 50\%), partially shadowed (between 50\% and 75\%), and overshadowed (above 75\%).}
\begin{tabular}{l|cccc}
& \multicolumn{4}{c}{\textbf{Mean shadow accumulation value}}
\\ \hline
\textbf{Season} & \multicolumn{1}{c|}{$<$ 25\%} & \multicolumn{1}{l|}{[25\%,50\%]} & \multicolumn{1}{c|}{[50\%,75\%]} & $>$ 75\% \\ \hline
Summer          & \multicolumn{1}{c|}{66.6\%}     & \multicolumn{1}{c|}{30.7\%}       & \multicolumn{1}{c|}{2.4\%}        & 0.3\%      \\
Spring          & \multicolumn{1}{c|}{61.7\%}     & \multicolumn{1}{c|}{25.5\%}       & \multicolumn{1}{c|}{9.2\%}        & 3.4\%      \\
Winter          & \multicolumn{1}{c|}{32.1\%}     & \multicolumn{1}{c|}{23.9\%}       & \multicolumn{1}{c|}{19.0\%}       & 25.0\%\\
\end{tabular}
\label{tab:dist_shadow}
\end{table}

\subsection{Sunlight access in Parisian parks}
Health researchers have revealed that access to sunlight in winter is as essential as shade in summer. A significant percentage of the population suffers vitamin D deficiency at the end of winter and consequently becomes vulnerable to diseases, including diabetes, heart disease, and osteoporosis~\cite{holick2004sunlight}. With the rapid growth in urbanization, sunlight access during winter within high-density municipal areas has been deemed critical more than ever~\cite{bethLetain, marielleMondon2015, hodyl, natalie2021}. However, people living or working in high-density environments, more often than not, have limited access to private green open spaces. And as development intensifies, the overshadowing of existing parks increases, creating tension between urban growth and the preservation of sunlight access~\cite{hodyl}.

In this case study, we investigated sunlight access in parks of Paris during winter. Table~\ref{tab:dist_shadow} shows the distribution of shadows in parks during the three seasons. Throughout winter, 25\% of the parks in Paris are overshadowed, i.e., they have a mean shadow accumulation value above 75\%. Also, there is a discrepancy in sunlight access comparing the distributions over the three seasons. The overshadowing of open urban spaces in Paris can largely be attributed to its densely built environment and narrow alleyways~\cite{nytimes2003, bloomberg2016}.

\begin{figure}[t]
  \centering
  \includegraphics[width=\linewidth]{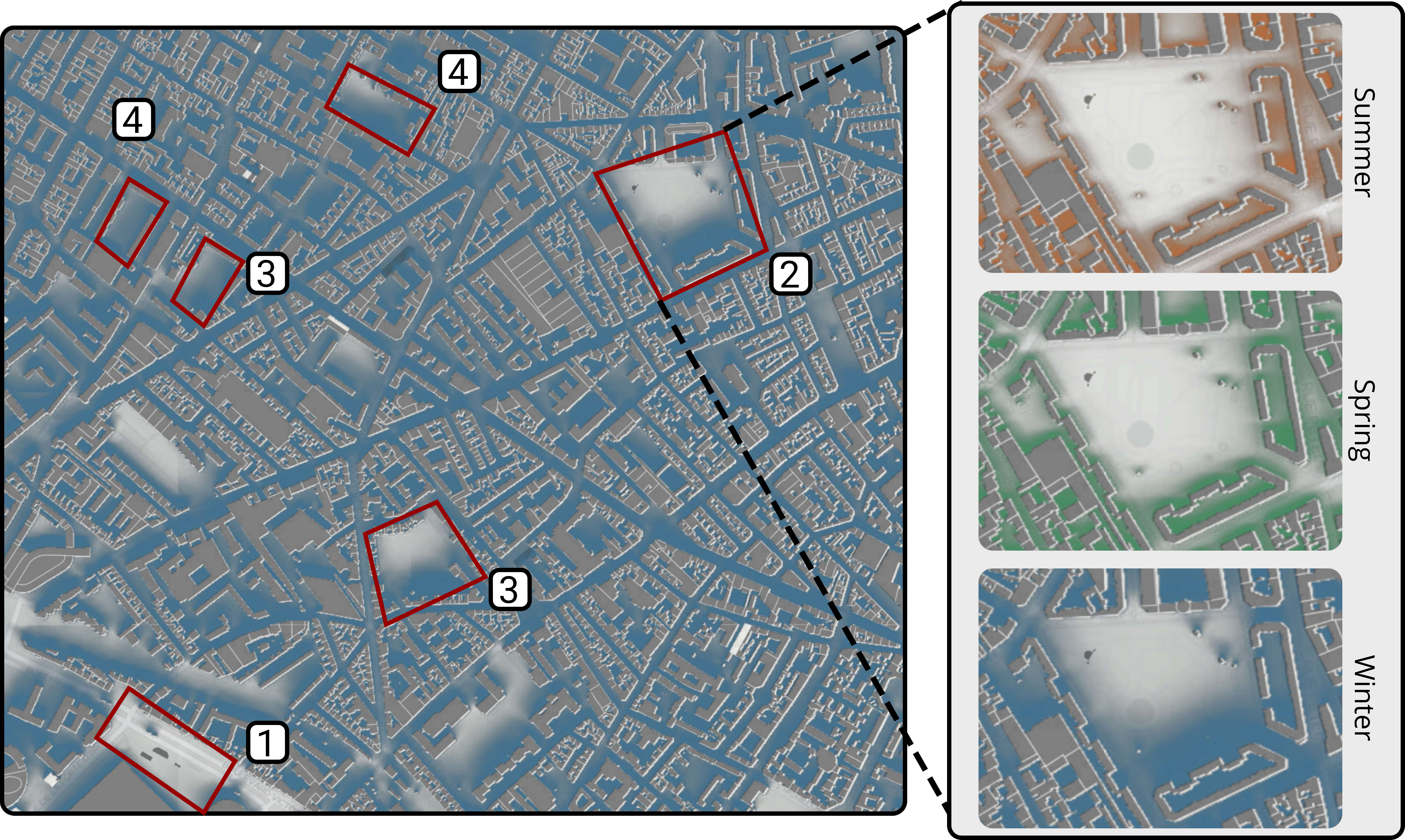}
  \caption{Visualizing accumulated shadow cast over parks in Paris. Left image highlights six parks divided into four categories of sunlight access during winter: (1) high sunlight access, (2) moderate sunlight access, (3) partial overshadowing and (4) significant overshadowing. Right images highlight shadow accumulation for one park (Square Saint Lambert).}
  \label{fig:parks}
\end{figure}

\begin{figure}[t]
  \centering
  \includegraphics[width=0.9\linewidth]{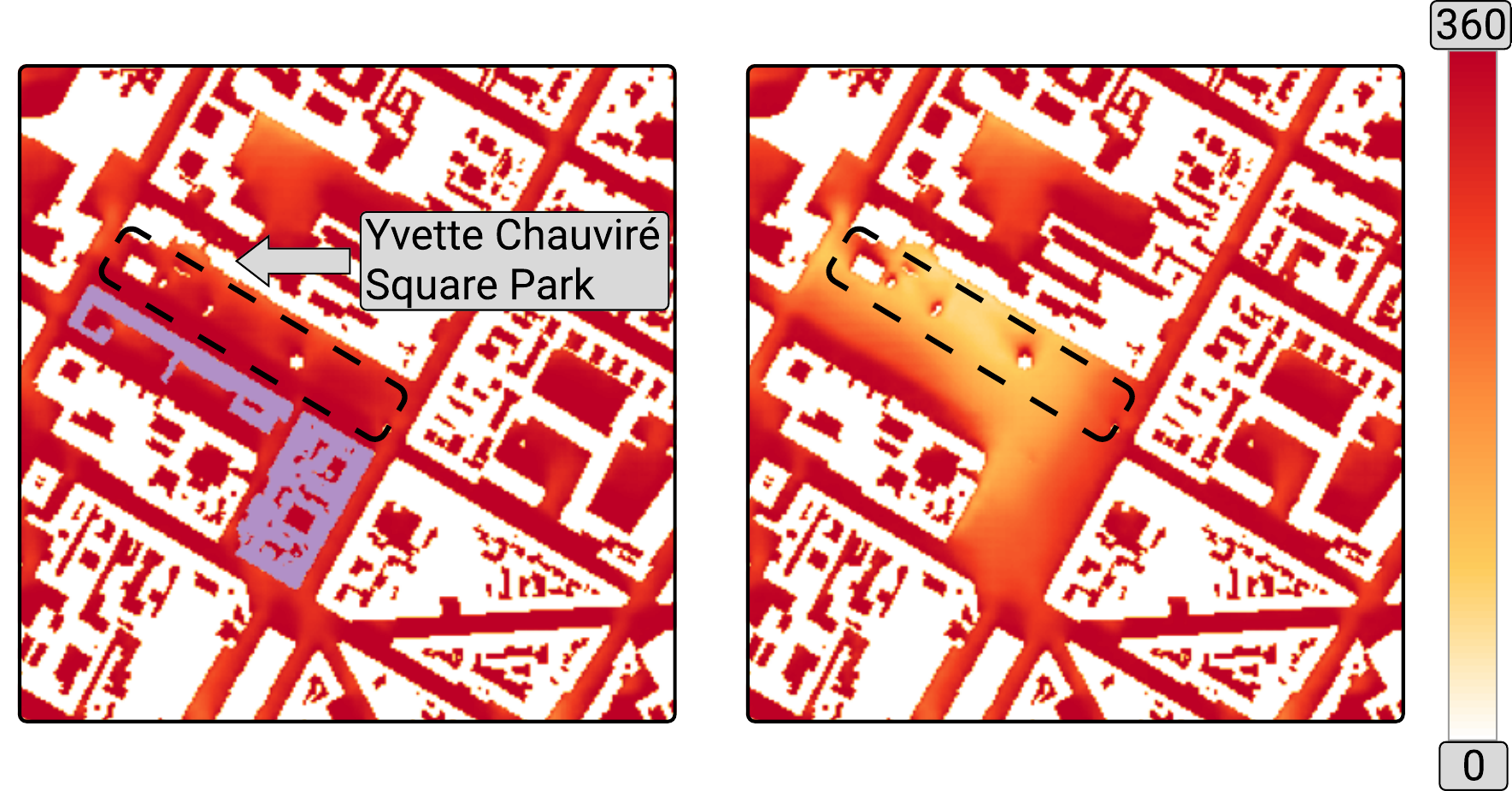}
  \vspace{-0.25cm}
  \caption{Comparison of the impact of nearby buildings on the overshadowing of Yvette Chauvir\'e Square Park (left) with an alternate scenario of without the buildings (right). The buildings considered are highlighted in purple. Removing these buildings too close to the park improves sunlight access by a significant amount.}
  \label{fig:parks_what_if}
\end{figure}

\begin{figure*}[t]
  \centering
  \includegraphics[width=1\linewidth]{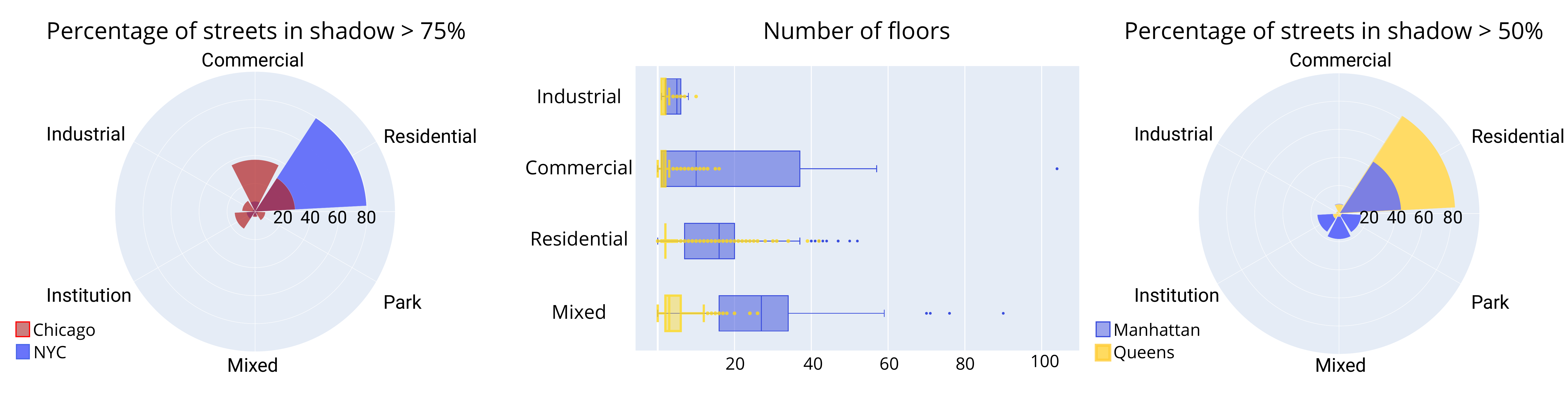}
  \caption{The percentage of streets in overshadowed areas per land use for Chicago and NYC during the winter (left). The distribution of building heights (middle) and the percentage of streets that are partially or overshadowed in Manhattan and Queens (right).}
  \label{fig:nyc}
\end{figure*}

We further looked into six public parks in the district of Vaugirard and divided them into four categories based on their mean shadow accumulation percentile score. The categories are: high sunlight access (below 25\%), moderate sunlight access (between 25\% and 50\%), partially shadowed (between 50\% and 75\%), and overshadowed (above 75\%). Figure~\ref{fig:parks} visualizes shadow accumulation during winter for these parks using our data.
Among them, two parks are partially shadowed, namely, Square du Clos Feuqui\`eres (59\%), and Square Duranton (70\%). Only Jardin \'Elisabeth Boselli has high sunlight access (22\%), whereas Square Saint-Lambert has moderate sunlight access (50\%). Conversely, Jardin Marguerite Boucicaut (78\%) and Square Yvette Chauvir\'e (89\%) lie in the overshadowed category.
%


To inspect the impact of nearby buildings on the overshadowing of Square Yvette Chauvir\'e, we performed a what-if scenario analysis by removing buildings too close to the park.
Figure~\ref{fig:parks_what_if} shows both the removed buildings~(left) and the resulting shadows~(right), highlighting \name's ability to assist in urban planning activities.
%
%
\name can help policy makers and government agencies in several ways. They can better identify locations for new parks within the municipality, 
and introduce sunlight protection policies considering new development scenarios.


\subsection{Analyzing land use-level street shadows}



The right to access sunlight and the impact of tall buildings on restricting this access in dense urban areas have long been the subject of popular debates~\cite{ bui2016mapping,seelye2017building}. Two main contributing factors to shadows on streets are street width and the dimension (i.e., height and bulk) of the surrounding buildings~\cite{andreou2014effect, zhu2020effect}. 
To investigate how the sunlight access pattern changes with land use in Chicago and NYC, in this case study we first looked at the average amount of accumulated shadow during the winter.~\highlight{Since we are interested in the impact of building morphology on shadow accumulation, we created street-level shadow data by aggregating shadow values at a street-segment level, using data from OSM~\cite{OpenStreetMap}. Each street segment considers a buffer of 5 meters, and we computed the mean accumulated shadow per segment by averaging pixel values within the buffered region. For each segment, we then assigned a land use type by spatially joining buffered streets with official land use datasets for Chicago~\cite{chicagolu} and NYC~\cite{pluto}.}
The results show no significant difference between the average shadow score in different land uses.
NYC has, on average, a slightly higher mean accumulated shadow, the maximum being 94\% in residential areas compared to 89\% for the same land use in Chicago. The distributions were also close, which does not reveal any specific pattern. 

Next, we computed the percentage of streets falling in the overshadowed range (above 75\%) for each land use type in both cities. 
Here, we see a more drastic difference. 
Although in NYC, mixed-use areas generally have a higher average of shadow score compared with residential ones, as Figure~\ref{fig:nyc}(left) shows, residential streets comprise 78\% of the total number of streets that are overshadowed during the winter. However, in Chicago, distribution is more balanced: overshadowed streets are primarily concentrated in commercial areas (39\% of the total number of streets), while residential areas comprise 25.5\% of total overshadowed streets. Streets around parks and open spaces in Chicago receive less sunlight than those in NYC (7\% vs. 2\%).

Concerns around equitable sunlight access age back to 1916, when NYC's first zoning resolution was adopted. To provide equitable access to sunlight, the regulations enforce restrictions on the height and bulk of the buildings. In NYC, the Queens borough's zoning regulations restrict it to mainly up to eight-story residential buildings, while in Manhattan tall skyscrapers dominate dense residential and mixed-use areas. \highlight{But do height restrictions in Queens provide residents with more sunlight access?}

In Queens, the mean accumulated shadow value in residential streets is 82\%, with a maximum of 96\%, while in Manhattan, we observed a lower average of 79\% and a lower maximum of 91\%. We looked at the percentages of streets falling within our ranges (Figure~\ref{fig:nyc}~(right)). \highlight{The publicly available PLUTO dataset provides information about the height and bulk of buildings across NYC~\cite{pluto}.} Using that data, we computed the distribution of building height per land use for each borough to have a more vivid picture and a fair comparison. Figure~\ref{fig:nyc}~(middle) shows that buildings in Manhattan have a much wider height range across all relevant land uses (we excluded parks and industrial areas from this analysis since they have very different zoning regulations) and generally are much taller than their counterparts in Queens. 
However, the radial plot (right) reveals an interesting pattern: although Manhattan buildings are much taller than Queens, 83\% of the shadowed streets of Queens are placed in residential areas, while this number drops to 43\% in Manhattan during the winter. This pattern, in huge part, is due to the bulky residential buildings and row houses built densely in Queens, which as the analysis shows, cast more shadows than the taller but skinnier buildings -- a pattern backed up by previous studies~\cite{miranda2018shadow}. 

Using our computed shadow data and the publicly available land use and building datasets, we could uncover an interesting and unexpected pattern in the Queens borough of NYC, which raises questions regarding the equitable access to sunlight and bulk and height regulations already in place. This analysis can be further replicated and extended to inform policy makers and residents in other cities about the built environment's impact on their daily lives.

\section{Conclusion and Future Work}
\label{sec:conclusion}

In this paper, we presented \name, a novel framework that enables the quantification of shadow and sunlight access in urban environments on a global scale.
We leveraged a conditional generative adversarial network to enable the real-time computation of OSM-like map tiles with shadow accumulation. \highlight{For a city such as NYC, it only takes 105 seconds to generate shadow tiles for the entire city, making it a viable and valuable solution for large-scale urban shadow and sunlight analysis.}

We also reported a series of experiments demonstrating the effectiveness of our approach, particularly taking into account cities from different continents. 
%
%
Due to the data training variety, the model can be used for  different seasons in any city of the globe (given available building height data). The model was evaluated using root mean square error, mean squared error, and mean absolute error, also considering the transferability to cities outside the training domain.
Moreover, we have made available the \dataset, a comprehensive dataset covering more than 100 cities over six continents. By doing this, we hope to facilitate research in multiple domains interested in shadow and sunlight access computation and alleviate data acquisition friction to benefit city planning activities as well as facilitate the development of concrete applications and studies that have shadows in cities as its object of analysis.

From an urban domain perspective, integrating advanced shadow-casting tools, such as \name, is fundamental in climate-resilient urban planning. The dynamic interplay of shadows, contingent upon the sun's seasonal path, significantly influences the energy dynamics and thermal comfort within urban environments. During the winter months, these tools empower policymakers to optimize access to sunlight, thereby mitigating energy consumption and enhancing urban comfort. Conversely, precise shadow calculations foster thermal comfort during summer by strategically shading pedestrian pathways and outdoor spaces. 
This holistic approach to shadow management is a cornerstone for resilient urban planning and design. 

However, there are still some unaddressed challenges and limitations that we plan to tackle in future work.
%
%
%
%
First, even though the amount of building height data available through OSM is extensive, several cities in the developing world lack good coverage. Therefore, we plan to explore alternatives such as utilizing widely available data (such as remote sensing) to enhance currently available height data.
Second, \name does not consider urban greenery, which impacts urban climate. Deciduous trees, for instance, offer a dual advantage, permitting sunlight penetration in winter while shielding against direct solar radiation in summer. In future work, we plan to extend our framework to also take into account trees.
\highlight{Third, while the findings of this study offer valuable insights, they should be interpreted with prudence when applied to policy formation or decision-making at the city level. Given the complexities and dynamic nature of urban environments, results may not fully encapsulate every variable and scenario. Therefore, we advise that our results and model  be used as a guiding framework rather than definitive solutions, encouraging further investigation and adaptation to specific urban contexts.}
%
%
%
%
Lastly, we plan to incorporate \name into GIS tools, such as QGIS. This will open new opportunities for studies in urban accessibility, targeting the identification of black ice, and urban heat island effect.

\ifCLASSOPTIONcaptionsoff
  \newpage
\fi



%
\bibliographystyle{IEEEtran}
\bibliography{references}

\begin{thebibliography}{10}
\providecommand{\url}[1]{#1}
\csname url@samestyle\endcsname
\providecommand{\newblock}{\relax}
\providecommand{\bibinfo}[2]{#2}
\providecommand{\BIBentrySTDinterwordspacing}{\spaceskip=0pt\relax}
\providecommand{\BIBentryALTinterwordstretchfactor}{4}
\providecommand{\BIBentryALTinterwordspacing}{\spaceskip=\fontdimen2\font plus
\BIBentryALTinterwordstretchfactor\fontdimen3\font minus
  \fontdimen4\font\relax}
\providecommand{\BIBforeignlanguage}[2]{{%
\expandafter\ifx\csname l@#1\endcsname\relax
\typeout{** WARNING: IEEEtran.bst: No hyphenation pattern has been}%
\typeout{** loaded for the language `#1'. Using the pattern for}%
\typeout{** the default language instead.}%
\else
\language=\csname l@#1\endcsname
\fi
#2}}
\providecommand{\BIBdecl}{\relax}
\BIBdecl

\bibitem{UN2019}
\BIBentryALTinterwordspacing
{UN Department of Economic and Social Affairs}. (2019) World urbanization
  prospects 2018: Highlights. [Online]. Available:
  \url{https://population.un.org/wup/Publications/Files/WUP2018-Highlights.pdf}
\BIBentrySTDinterwordspacing

\bibitem{ding2015inclusive}
X.~Ding, W.~Zhong, R.~G. Shearmur, X.~Zhang, and D.~Huisingh, ``An inclusive
  model for assessing the sustainability of cities in developing
  countries--trinity of cities' sustainability from spatial, logical and time
  dimensions (tcs-sltd),'' \emph{Journal of Cleaner Production}, vol. 109, pp.
  62--75, 2015.

\bibitem{jasonMBarr19}
\BIBentryALTinterwordspacing
J.~M. Barr. (2019) Skyscrapers and shadows: The value of sunshine in the city.
  [Online]. Available: \url{https://buildingtheskyline.org/city-shadows}
\BIBentrySTDinterwordspacing

\bibitem{miranda2018shadow}
F.~Miranda, H.~Doraiswamy, M.~Lage, L.~Wilson, M.~Hsieh, and C.~T. Silva,
  ``Shadow accrual maps: Efficient accumulation of city-scale shadows over
  time,'' \emph{IEEE Transactions on Visualization and Computer Graphics},
  vol.~25, no.~3, pp. 1559--1574, 2018.

\bibitem{compagnon2004solar}
R.~Compagnon, ``Solar and daylight availability in the urban fabric,''
  \emph{Energy and buildings}, vol.~36, no.~4, pp. 321--328, 2004.

\bibitem{kruger2011impact}
E.~Kr{\"u}ger, F.~Minella, and F.~Rasia, ``Impact of urban geometry on outdoor
  thermal comfort and air quality from field measurements in {Curitiba,
  Brazil},'' \emph{Building and Environment}, vol.~46, no.~3, pp. 621--634,
  2011.

\bibitem{jiang2017sunchase}
L.~Jiang, Y.~Hua, C.~Ma, and X.~Liu, ``Sunchase: Energy-efficient route
  planning for solar-powered evs,'' in \emph{2017 IEEE 37th International
  Conference on Distributed Computing Systems (ICDCS)}.\hskip 1em plus 0.5em
  minus 0.4em\relax IEEE, 2017, pp. 383--393.

\bibitem{vulkan2018modeling}
A.~Vulkan, I.~Kloog, M.~Dorman, and E.~Erell, ``Modeling the potential for pv
  installation in residential buildings in dense urban areas,'' \emph{Energy
  and Buildings}, vol. 169, pp. 97--109, 2018.

\bibitem{knowles1980solar}
R.~L. Knowles, ``The solar envelope,'' \emph{Solar L. Rep.}, vol.~2, p. 263,
  1980.

\bibitem{de2019novel}
F.~De~Luca and T.~Dogan, ``A novel solar envelope method based on solar
  ordinances for urban planning,'' in \emph{Building Simulation}, vol.~12,
  no.~5.\hskip 1em plus 0.5em minus 0.4em\relax Springer, 2019, pp. 817--834.

\bibitem{alzoubi2010low}
H.~H. Alzoubi and A.~A. Alshboul, ``Low energy architecture and solar rights:
  Restructuring urban regulations, view from jordan,'' \emph{Renewable Energy},
  vol.~35, no.~2, pp. 333--342, 2010.

\bibitem{pereira2001methodology}
F.~O.~R. Pereira, C.~A.~N. Silva, and B.~Turkienikz, ``A methodology for
  sunlight urban planning: a computer-based solar and sky vault obstruction
  analysis,'' \emph{Solar energy}, vol.~70, no.~3, pp. 217--226, 2001.

\bibitem{hachem2013solar}
C.~Hachem, P.~Fazio, and A.~Athienitis, ``Solar optimized residential
  neighborhoods: Evaluation and design methodology,'' \emph{Solar Energy},
  vol.~95, pp. 42--64, 2013.

\bibitem{vartholomaios2015residential}
A.~Vartholomaios, ``The residential solar block envelope: A method for enabling
  the development of compact urban blocks with high passive solar potential,''
  \emph{Energy and Buildings}, vol.~99, pp. 303--312, 2015.

\bibitem{kanters2016planning}
J.~Kanters and M.~Wall, ``A planning process map for solar buildings in urban
  environments,'' \emph{Renewable and Sustainable Energy Reviews}, vol.~57, pp.
  173--185, 2016.

\bibitem{andreou2014effect}
E.~Andreou, ``The effect of urban layout, street geometry and orientation on
  shading conditions in urban canyons in the mediterranean,'' \emph{Renewable
  Energy}, vol.~63, pp. 587--596, 2014.

\bibitem{deng2020impact}
J.-Y. Deng and N.~H. Wong, ``Impact of urban canyon geometries on outdoor
  thermal comfort in central business districts,'' \emph{Sustainable Cities and
  Society}, vol.~53, p. 101966, 2020.

\bibitem{seelye2017building}
K.~Q. Seelye, ``Building boom in {Boston} casts shadows on history and public
  space,'' \emph{The New York Times}, vol.~11, 2017.

\bibitem{costamagna2019livability}
F.~Costamagna, R.~Lind, and O.~Stjernstr{\"o}m, ``Livability of urban public
  spaces in northern {Swedish} cities: The case of {Ume{\aa}},'' \emph{Planning
  Practice \& Research}, vol.~34, no.~2, pp. 131--148, 2019.

\bibitem{zhu2019solar}
R.~Zhu, L.~You, P.~Santi, M.~S. Wong, and C.~Ratti, ``Solar accessibility in
  developing cities: A case study in kowloon east, hong kong,''
  \emph{Sustainable Cities and Society}, vol.~51, p. 101738, 2019.

\bibitem{eisemann2011real}
E.~Eisemann, M.~Schwarz, U.~Assarsson, and M.~Wimmer, \emph{Real-time
  shadows}.\hskip 1em plus 0.5em minus 0.4em\relax CRC Press, 2011.

\bibitem{nah2014sato}
J.-H. Nah and D.~Manocha, ``Sato: Surface area traversal order for shadow ray
  tracing,'' in \emph{Computer Graphics Forum}, vol.~33, no.~6.\hskip 1em plus
  0.5em minus 0.4em\relax Wiley Online Library, 2014, pp. 167--177.

\bibitem{zhang2019shadowgan}
S.~Zhang, R.~Liang, and M.~Wang, ``Shadowgan: Shadow synthesis for virtual
  objects with conditional adversarial networks,'' \emph{Computational Visual
  Media}, vol.~5, no.~1, pp. 105--115, 2019.

\bibitem{liu2020arshadowgan}
D.~Liu, C.~Long, H.~Zhang, H.~Yu, X.~Dong, and C.~Xiao, ``Arshadowgan: Shadow
  generative adversarial network for augmented reality in single light
  scenes,'' in \emph{Proceedings of the IEEE/CVF Conference on Computer Vision
  and Pattern Recognition}, 2020, pp. 8139--8148.

\bibitem{tan2021urban}
J.~K. Tan, R.~N. Belcher, H.~T. Tan, S.~Menz, and T.~Schroepfer, ``The urban
  heat island mitigation potential of vegetation depends on local surface type
  and shade,'' \emph{Urban Forestry \& Urban Greening}, vol.~62, p. 127128,
  2021.

\bibitem{hodyl}
\BIBentryALTinterwordspacing
Hodyl and Co. (2018) Central {Melbourne} sunlight report. [Online]. Available:
  \url{https://www.hodyl.co/projects/sunlight-to-public-open-space}
\BIBentrySTDinterwordspacing

\bibitem{shadeMap}
\BIBentryALTinterwordspacing
Shademap: the definitive source of shade data for the planet. [Online].
  Available: \url{https://shademap.app/}
\BIBentrySTDinterwordspacing

\bibitem{jveuxDuSoleil}
\BIBentryALTinterwordspacing
{JveuxDuSoleil}: a web-app to simulate the shadow of cities and detect sunny
  terraces. [Online]. Available: \url{https://jveuxdusoleil.fr}
\BIBentrySTDinterwordspacing

\bibitem{arcgis}
\BIBentryALTinterwordspacing
{ArcGIS Pro}. [Online]. Available:
  \url{https://www.esri.com/en-us/arcgis/products/arcgis-pro/overview}
\BIBentrySTDinterwordspacing

\bibitem{qgis}
\BIBentryALTinterwordspacing
{QGIS}: a free and open source geographic information system. [Online].
  Available: \url{https://qgis.org/en/site/}
\BIBentrySTDinterwordspacing

\bibitem{solarApiGoogle}
\BIBentryALTinterwordspacing
{Google Solar API}. [Online]. Available:
  \url{https://developers.google.com/maps/documentation/solar}
\BIBentrySTDinterwordspacing

\bibitem{angelidou2019social}
M.~Angelidou and A.~Psaltoglou, ``Social innovation, games and urban planning:
  An analysis of current approaches,'' \emph{International Journal of
  Electronic Governance}, vol.~11, no.~1, pp. 5--22, 2019.

\bibitem{kavouras2023low}
I.~Kavouras, E.~Sardis, E.~Protopapadakis, I.~Rallis, A.~Doulamis, and
  N.~Doulamis, ``A low-cost gamified urban planning methodology enhanced with
  co-creation and participatory approaches,'' \emph{Sustainability}, vol.~15,
  no.~3, p. 2297, 2023.

\bibitem{moreira2023urban}
G.~Moreira, M.~Hosseini, M.~N.~A. Nipu, M.~Lage, N.~Ferreira, and F.~Miranda,
  ``{The Urban Toolkit}: A grammar-based framework for urban visual
  analytics,'' \emph{IEEE Transactions on Visualization and Computer Graphics},
  2024.

\bibitem{yap2023urbanity}
W.~Yap, R.~Stouffs, and F.~Biljecki, ``Urbanity: Automated modeling and
  analysis of multidimensional networks in cities,'' \emph{npj Urban
  Sustainability}, vol.~3, no.~1, p.~45, 2023.

\bibitem{yap2022free}
W.~Yap, P.~Janssen, and F.~Biljecki, ``Free and open source urbanism: Software
  for urban planning practice,'' \emph{Computers, Environment and Urban
  Systems}, vol.~96, p. 101825, 2022.

\bibitem{10.1145/3394486.3403127}
Y.~Zhang, Y.~Li, X.~Zhou, X.~Kong, and J.~Luo, ``Curb-gan: Conditional urban
  traffic estimation through spatio-temporal generative adversarial networks,''
  in \emph{Proceedings of the 26th ACM SIGKDD International Conference on
  Knowledge Discovery \& Data Mining}, 2020, p. 842–852.

\bibitem{isola2017image}
P.~Isola, J.-Y. Zhu, T.~Zhou, and A.~A. Efros, ``Image-to-image translation
  with conditional adversarial networks,'' in \emph{Proceedings of the IEEE
  conference on Computer Vision and Pattern Recognition}, 2017, pp. 1125--1134.

\bibitem{zheng2016visual}
Y.~Zheng, W.~Wu, Y.~Chen, H.~Qu, and L.~M. Ni, ``Visual analytics in urban
  computing: An overview,'' \emph{IEEE Transactions on Big Data}, vol.~2,
  no.~3, pp. 276--296, 2016.

\bibitem{10.1145/3397536.3422268}
D.~Wang, Y.~Fu, P.~Wang, B.~Huang, and C.-T. Lu, ``Reimagining city
  configuration: Automated urban planning via adversarial learning,'' in
  \emph{Proceedings of the 28th International Conference on Advances in
  Geographic Information Systems}, 2020, p. 497–506.

\bibitem{mokhtar2020conditional}
S.~Mokhtar, A.~Sojka, and C.~C. Davila, ``Conditional generative adversarial
  networks for pedestrian wind flow approximation,'' in \emph{Proceedings of
  the 11th Annual Symposium on Simulation for Architecture and Urban Design},
  2020, pp. 1--8.

\bibitem{zhang2019trafficgan}
Y.~Zhang, Y.~Li, X.~Zhou, X.~Kong, and J.~Luo, ``{TrafficGAN}: Off-deployment
  traffic estimation with traffic generative adversarial networks,'' in
  \emph{2019 IEEE International Conference on Data Mining (ICDM)}, 2019, pp.
  1474--1479.

\bibitem{zhang2020cgail}
X.~Zhang, Y.~Li, X.~Zhou, and J.~Luo, ``{cGAIL}: Conditional generative
  adversarial imitation learning—an application in taxi drivers’ strategy
  learning,'' \emph{IEEE Transactions on Big Data}, vol.~8, no.~5, pp.
  1288--1300, 2020.

\bibitem{yuan2022stgan}
Y.~Yuan, Y.~Zhang, B.~Wang, Y.~Peng, Y.~Hu, and B.~Yin, ``{STGAN}:
  Spatio-temporal generative adversarial network for traffic data imputation,''
  \emph{IEEE Transactions on Big Data}, vol.~9, no.~1, pp. 200--211, 2022.

\bibitem{wu2022generative}
A.~N. Wu, R.~Stouffs, and F.~Biljecki, ``Generative adversarial networks in the
  built environment: A comprehensive review of the application of gans across
  data types and scales,'' \emph{Building and Environment}, p. 109477, 2022.

\bibitem{OpenStreetMap}
\BIBentryALTinterwordspacing
{OpenStreetMap}. [Online]. Available: \url{https://www.openstreetmap.org}
\BIBentrySTDinterwordspacing

\bibitem{central_park}
\BIBentryALTinterwordspacing
{Central Park Sunshine Task Force Committee}. (2015) {Central Park} sunshine
  taskforce report. [Online]. Available:
  \url{https://www.cb5.org/cb5m/resolutions/2015-may/may-2015_10}
\BIBentrySTDinterwordspacing

\bibitem{bui2016mapping}
\BIBentryALTinterwordspacing
Q.~Bui and J.~White. (2016) Mapping the shadows of {New York City}: Every
  building every block. [Online]. Available:
  \url{https://www.nytimes.com/interactive/2016/12/21/upshot/Mapping-the-Shadows-of-New-York-City.html}
\BIBentrySTDinterwordspacing

\bibitem{mirza2014conditional}
M.~Mirza and S.~Osindero, ``Conditional generative adversarial nets,''
  \emph{arXiv preprint arXiv:1411.1784}, 2014.

\bibitem{krizhevsky2012imagenet}
A.~Krizhevsky, I.~Sutskever, and G.~E. Hinton, ``Imagenet classification with
  deep convolutional neural networks,'' \emph{Advances in neural information
  processing systems}, vol.~25, 2012.

\bibitem{pathak2016context}
D.~Pathak, P.~Krahenbuhl, J.~Donahue, T.~Darrell, and A.~A. Efros, ``Context
  encoders: Feature learning by inpainting,'' in \emph{Proceedings of the IEEE
  conference on computer vision and pattern recognition}, 2016, pp. 2536--2544.

\bibitem{wang2004image}
Z.~Wang, A.~C. Bovik, H.~R. Sheikh, and E.~P. Simoncelli, ``Image quality
  assessment: from error visibility to structural similarity,'' \emph{IEEE
  Transactions on Image Processing}, vol.~13, no.~4, pp. 600--612, 2004.

\bibitem{paul2022edge}
S.~Paul, B.~Jhamb, D.~Mishra, and M.~S. Kumar, ``Edge loss functions for
  deep-learning depth-map,'' \emph{Machine Learning with Applications}, vol.~7,
  p. 100218, 2022.

\bibitem{wang2022esa}
L.~Wang, L.~Wang, and S.~Chen, ``{ESA-CycleGAN}: Edge feature and
  self-attention based cycle-consistent generative adversarial network for
  style transfer,'' \emph{IET Image Processing}, vol.~16, no.~1, pp. 176--190,
  2022.

\bibitem{wang2018high}
T.-C. Wang, M.-Y. Liu, J.-Y. Zhu, A.~Tao, J.~Kautz, and B.~Catanzaro,
  ``High-resolution image synthesis and semantic manipulation with conditional
  {GANs},'' in \emph{Proceedings of the IEEE conference on computer vision and
  pattern recognition}, 2018, pp. 8798--8807.

\bibitem{ronneberger2015u}
O.~Ronneberger, P.~Fischer, and T.~Brox, ``U-net: Convolutional networks for
  biomedical image segmentation,'' in \emph{International Conference on Medical
  Image Computing and Computer-assisted Intervention}.\hskip 1em plus 0.5em
  minus 0.4em\relax Springer, 2015, pp. 234--241.

\bibitem{jian2016deep}
S.~Jian, H.~Kaiming, R.~Shaoqing, and Z.~Xiangyu, ``Deep residual learning for
  image recognition,'' in \emph{IEEE Conference on Computer Vision \& Pattern
  Recognition}, 2016, pp. 770--778.

\bibitem{hu2018image}
J.~Hu, W.~Yu, and Z.~Yu, ``Image-to-image translation with conditional-{GAN},''
  \emph{CS230: Deep Learning, Spring}, 2018.

\bibitem{hosseini_towards_2022}
M.~Hosseini, M.~Saugstad, F.~Miranda, A.~Sevtsuk, C.~Silva, and J.~Froehlich,
  ``Towards global-scale crowd+{AI} techniques to map and assess sidewalks for
  people with disabilities,'' in \emph{Proceedings of the CVPR 2022 AVA
  (Accessibility, Vision, and Autonomy Meet) Workshop}, 2022.

\bibitem{froehlich_future_2022}
J.~E. Froehlich, Y.~Eisenberg, M.~Hosseini, F.~Miranda \emph{et~al.}, ``The
  future of urban accessibility for people with disabilities: Data collection,
  analytics, policy, and tools,'' in \emph{Proceedings of the 24th
  International ACM SIGACCESS Conference on Computers and Accessibility}, 2022.

\bibitem{hosseini_mapping_2023}
M.~Hosseini, A.~Sevtsuk, F.~Miranda, R.~M. Cesar~Jr, and C.~T. Silva, ``Mapping
  the walk: A scalable computer vision approach for generating sidewalk network
  datasets from aerial imagery,'' \emph{Computers, Environment and Urban
  Systems}, vol. 101, p. 101950, 2023.

\bibitem{liu_effects_2014}
Z.~Liu and J.~Heer, ``The effects of interactive latency on exploratory visual
  analysis,'' \emph{IEEE Transactions on Visualization and Computer Graphics},
  vol.~20, no.~12, pp. 2122--2131, 2014.

\bibitem{holick2004sunlight}
M.~F. Holick, ``Sunlight and vitamin {D} for bone health and prevention of
  autoimmune diseases, cancers, and cardiovascular disease,'' \emph{The
  American journal of clinical nutrition}, vol.~80, no.~6, pp. 1678S--1688S,
  2004.

\bibitem{bethLetain}
\BIBentryALTinterwordspacing
{The Parks Council, NYC}. (1991) Preserving sunlight in {New York City}'s
  parks: A zoning proposal. [Online]. Available:
  \url{https://beth-letain.squarespace.com/s/Preserving-Sunshine-Zoning.pdf}
\BIBentrySTDinterwordspacing

\bibitem{marielleMondon2015}
\BIBentryALTinterwordspacing
M.~Mondon. (2015) {NYC} councilman wants to preserve {Central Park} sunshine.
  [Online]. Available:
  \url{https://nextcity.org/urbanist-news/central-park-sunshine-high-rise-shadows-new-committee}
\BIBentrySTDinterwordspacing

\bibitem{natalie2021}
\BIBentryALTinterwordspacing
N.~Filatoff. (2021) Let there be sunlight ... in the public parks of
  high-density communities! [Online]. Available:
  \url{https://www.pv-magazine-australia.com/2021/09/20/let-there-be-sunlight-in-the-public-parks-of-high-density-communities}
\BIBentrySTDinterwordspacing

\bibitem{nytimes2003}
\BIBentryALTinterwordspacing
E.~Sciolino. (2003) Paris journal; call it the city of darkness, and give it
  vitamin {D}. [Online]. Available:
  \url{https://www.nytimes.com/2003/01/06/world/paris-journal-call-it-the-city-of-darkness-and-give-it-vitamin-d.html}
\BIBentrySTDinterwordspacing

\bibitem{bloomberg2016}
\BIBentryALTinterwordspacing
A.~Small. (2016) Why are {European} cities so dense? [Online]. Available:
  \url{https://www.bloomberg.com/news/articles/2016-10-27/why-european-cities-still-have-more-dense-development}
\BIBentrySTDinterwordspacing

\bibitem{zhu2020effect}
R.~Zhu, M.~S. Wong, L.~You, P.~Santi, J.~Nichol, H.~C. Ho, L.~Lu, and C.~Ratti,
  ``The effect of urban morphology on the solar capacity of three-dimensional
  cities,'' \emph{Renewable Energy}, vol. 153, pp. 1111--1126, 2020.

\bibitem{chicagolu}
\BIBentryALTinterwordspacing
{The Chicago Metropolitan Agency for Planning (CMAP)}. {Land Use Data}.
  [Online]. Available: \url{https://www.cmap.illinois.gov/data/land-use}
\BIBentrySTDinterwordspacing

\bibitem{pluto}
\BIBentryALTinterwordspacing
{New York City Department of City Planning (NYC DCP)}. {PLUTO and MapPLUTO}.
  [Online]. Available:
  \url{https://www1.nyc.gov/site/planning/data-maps/open-data/dwn-pluto-mappluto.page}
\BIBentrySTDinterwordspacing

\end{thebibliography}

\noindent \textbf{Kazi Shahrukh Omar} is a Computer Science PhD student at UIC, interested in urban computing and visualization.

\noindent \textbf{Gustavo Moreira} is a Computer Science PhD student at UIC, interested in visualization and visual analytics.

\noindent \textbf{Daniel Hodczak} is an undergraduate student at UIC, interested in designing frameworks for efficient urban research. 

\noindent \textbf{Maryam Hosseini} is a Postdoctoral Associate at the City Form Lab at MIT. She received the Ph.D. in Urban Systems from Rutgers University. She works in the intersection between visualization, computer vision and urban science.

\noindent \textbf{Nicola Colaninno} is an Assistant Professor at the Polytechnic University of Milan. He holds a Ph.D. in Urban and Architectural Mgmt. and Valuations from UPC. His research focuses on urban planning and urban climate analysis.

\noindent \textbf{Marcos Lage} is an Associate Professor in the Dept. of Computer Science at UFF. He received the Ph.D. in Applied Mathematics from PUC-Rio. His research interests include visual computing and topological data structures.

\noindent \textbf{Fabio Miranda} is an Assistant Professor in the Dept. of Computer Science at UIC. He received the Ph.D. in Computer Science from NYU. His research focuses on large-scale data analysis, data structures, and urban data visualization.












\end{document}